%% file: submission.tex
\documentclass[preprint,12pt,number, 1p]{elsarticle}

\usepackage{amssymb}
\usepackage{lineno}
\usepackage{pgfplots}
\usepackage{xcolor}
\usepgfplotslibrary{fillbetween}
\usepackage{subfigure}
\usepackage{algorithm2e}
\usepackage{verbatim}
\usepackage{multirow}

\journal{arXiv}
\date{}

\begin{document}

\begin{frontmatter}

\title{Assessing Streamline Plausibility Through Randomized Iterative Spherical-Deconvolution Informed Tractogram Filtering}

\author[inst1]{Antonia Hain}
\ead{s8anhain@stud.uni-saarland.de}

\affiliation[inst1]{organization={Saarland University, Faculty of Mathematics and Computer Science},
            addressline={Campus E1.7}, 
            city={Saarbruecken},
            postcode={66041}, 
            state={Saarland},
            country={Germany}}

\author[inst2,inst3]{Daniel J\"orgens}
\ead{danjorg@kth.se}
\author[inst3]{Rodrigo Moreno\corref{CorrespondingAuthor}}
\ead{rodmore@kth.se}

\affiliation[inst2]{organization={Division of Brain, Imaging, and Behaviour, Krembil Research Institute, Toronto Western Hospital, University Health Network},city={Toronto},state={Ontario},country={Canada}}

\affiliation[inst3]{organization={KTH Royal Institute of Technology, Department of Biomedical Engineering and Health Systems},
            addressline={Halsovagen 11C}, 
            city={Huddinge},
            postcode={14157}, 
            state={Stockholm},
            country={Sweden}}
            
\cortext[CorrespondingAuthor]{Corresponding author}

\begin{abstract}
Tractography has become an indispensable part of brain connectivity studies. However, it is currently facing problems with reliability. In particular, a substantial amount of nerve fiber reconstructions (\textit{streamlines}) in tractograms produced by state-of-the-art tractography methods are anatomically implausible. To address this problem, tractogram filtering methods have been developed to remove faulty connections in a postprocessing step. This study takes a closer look at one such method, \textit{Spherical-deconvolution Informed Filtering of Tractograms} (SIFT), which uses a global optimization approach to improve the agreement between the remaining streamlines after filtering and the underlying diffusion magnetic resonance imaging data. SIFT is not suitable to judge the plausibility of individual streamlines since its results depend on the size and composition of the surrounding tractogram. To tackle this problem, we propose applying SIFT to randomly selected tractogram subsets in order to retrieve multiple assessments for each streamline. This approach makes it possible to identify streamlines with very consistent filtering results, which were used as pseudo ground truths for training classifiers. The trained classifier is able to distinguish the obtained groups of plausible and implausible streamlines with accuracy above 80\%. 

\end{abstract}


\end{frontmatter}


\section{Introduction}
\label{sec:sample1}

Tractography uses data acquired with diffusion-weighted magnetic resonance imaging (DW-MRI) to trace nerve fiber tracts in the brain, producing a model called a \textit{tractogram}. 
A tractogram consists of a set of \textit{streamlines}, each of which represents an assumed nerve fiber through an ordered set of 3-dimensional coordinates \cite{Ugurlu:2019}. The main goal of tractography can be described as creating a set of streamlines that pose a maximally accurate, digital representation of the \textit{structural connectome} \cite{Jeurissen:2017}, which refers to the actual set of nerve fibers in the brain \cite{Frigo:2020}.
Applications for tractography include neurosurgery planning \cite{Panesar2019}, the study of neurological diseases \cite{Yamada:2009}, or the scientific study of the brain to understand its function and its links to human behavior \cite{Assaf:2017}.

By generating many streamlines (usually millions), current tractography methods have been shown to be able to recover all relevant bundles \cite{Rheault:2020}, at least if they are run in an appropriate way \cite{Schilling:2020,Schilling:2021}. However, they are also prone to generating erroneous and implausible streamlines. It has been estimated that an average of four false-positive streamlines are generated for each valid streamline in the tractogram \cite{Maier-Hein:2016}. This can hardly be fully avoided because tractography methods with high sensitivity tend to show low specificity and vice versa \cite{Thomas:2014}. In order to take steps towards a reliable model of the connectome, removing implausible connections is crucial. 

A whole field of research, commonly called \textit{tractogram filtering}, has been established as a way to deal with the excess of false-positive streamlines. Tractogram filtering approaches remove streamlines from tractograms by evaluating their plausibility, for example, by considering the geometrical properties of streamlines (e.g., \cite{ExTractor:2021}), anatomical constraints (e.g., \cite{Wassermann:2016}), through clustering approaches (e.g., \cite{RecoBundles:2018}) or by correspondence to the underlying DW-MRI data (e.g., \cite{SIFT:2012}). A review of methods is presented in \cite{Joergens:2021}. 

Identifying implausible streamlines can be posed as a binary classification problem: Each streamline is assigned a ``positive'' (P) or ``negative'' (N) label if it appears to be plausible or implausible, respectively. Streamlines with a ``negative'' label are subsequently removed from the tractogram. The terms ``true positive" (TP) and ``false positive'' (FP) frequently appear in tractography literature as well, even though they express notions that are slightly different from the  meaning used in statistics \cite{Jbabdi:2011}: Here, a ``true positive" streamline is regarded to be a fully correct reconstruction of an actual, plausible nerve fiber bundle, while a ``false positive'' streamline represents a faulty/noisy reconstruction that does not accurately represent any existing structure in the brain. Therefore, in this work, we use the terms ``plausible'' or ``implausible'' interchangeably with ``positive'' and ``negative'' or ``true positive" and ``false positive'', even though the concepts are not entirely equal in the classical sense.

One popular tractogram filtering method is \textit{Spherical-deconvolution Informed Filtering of Tractograms} (SIFT) \cite{SIFT:2012}. This approach belongs to the family of methods that assess tractogram quality by comparing the acquired DW-MRI data to the expected one from the tractogram \cite{Joergens:2021}. In particular, SIFT removes streamlines to increase the consistency of the tractogram with respect to the acquired data based on a global optimization approach. Other examples of such methods include LiFE \cite{LiFe:2014}, COMMIT \cite{COMMIT:2014}, SIFT2 \cite{Smith:2015a}, and COMMIT2 \cite{Schiavi:2020,Ocampo:2021,Sairanen:2022}.
SIFT does not operate directly on raw data, but instead uses the \textit{fiber orientation distribution} (FOD) \cite{Tournier:2004}. 
A tractogram can be filtered by investigating how the effect of removing an individual streamline manifests in the streamline density in each voxel compared to the corresponding local FOD along the streamline. Streamlines that increase the mismatch between streamline density and local FODs are prioritized for filtering. 
Given the complexity of the computation and the number of streamlines, trying to find a globally optimal solution is infeasible. Therefore, a gradient-descent approach is used to reduce a cost function. 

Several criteria for termination of the optimization process are provided by SIFT. By default, streamlines are removed until the cost function gradient of the candidate streamlines becomes sufficiently small. 
This option is referred to as filtering ``to convergence''. In the absence of other termination criteria which specify a maximum number of streamlines to be removed, filtering to convergence will naturally lead to the most reliable result set of streamlines, but with the streamline density being decreased the most. In fact, it has been argued that the remaining number of streamlines is not always sufficient for quantitative analyses, and for that reason, the authors of SIFT recommend applying the method to tractograms with a high number of streamlines \cite{SIFT2:2015}. 

Due to the design of SIFT's cost function, streamlines will be removed if the track density in their path is too high to match (parts of) the FODs. 
Such a mismatch could be due to different reasons, including at least the following cases:

\begin{enumerate}
\item The FOD representation in a voxel is not accurately describing the underlying anatomy. This can happen, for example, when the raw data is not able to model and resolve different fiber populations. 
\item The streamlines may be (partly) faulty/noisy. This is the target case for filtering, as the streamline is a false positive and must be removed.
\item The streamlines are plausible, but the streamline density in the corresponding voxel fractions is simply exaggerated. It is common for tractography methods to create multiple similar streamlines. 
We will refer to these streamlines as \textit{redundants}.
\end{enumerate}

Thus, a streamline being rejected by SIFT is therefore not a sufficient indicator of it being implausible. 
Since SIFT removes streamlines from these three categories, SIFT as such is not completely suitable for assessing the correspondence of an \textit{individual} streamline to the DW-MRI data. The focus of this paper is, therefore, to distinguish cases two and three.

Machine learning has been used for training classifiers to speed up the processing of expensive tractogram filtering methods \cite{Astolfi2020,Joergens:2022}. Although it is appealing to train a binary classifier to distinguish false-positive from true-positive streamlines based on the raw output of SIFT, this is unfortunately not possible for the abovementioned reasons. Indeed, redundant and true positive streamlines share similar features, but SIFT rejects the former and accepts the latter. Thus, for SIFT to be used in classifiers of streamlines, it is necessary to have a method to distinguish between false positives and redundants.

We have found that specific streamlines may be classified differently by SIFT depending on the composition of the tractogram. 
In this paper, we take advantage of this property with the goal of disentangling false positives from redundant streamlines. More specifically, we apply SIFT on randomized subsets of tractograms to identify both of those streamline groups.
Since SIFT seems to yield inconsistent results for certain streamlines when found in different subsets, we spotlight those streamlines and explore how they might be appropriately labeled. We refer to the proposed approach as \textit{randomized SIFT} (rSIFT).

\section{Materials and Methods}

\subsection{Datasets}
We carried out the experiments on pre-processed data \cite{Glasser:2013} of six different subjects from the young adult data set of the Human Connectome Project \cite{HCP:2013}. The whole-brain tractograms derived from this data were provided by the authors of \cite{TractSeg:2018}\footnote{https://doi.org/10.5281/zenodo.1088277} and computed with the iFOD2 \cite{Tournier:2010} algorithm. Ten million streamlines were created for each subject, restricted to be between 40mm and 250mm in length, with a step size of 0.625mm. The tracking was further constrained by anatomical priors based on the segmentation of different tissue types in the brain \cite{Smith:2012ACT}.
In order to make computation more feasible, we used an additional post-processing step in which the streamlines were compressed to smaller sets of coordinates with the method in  \cite{Presseau:2015} using a tolerance error of 0.35mm.

\begin{figure}

\centering
\small

\includegraphics[width=0.7\textwidth]{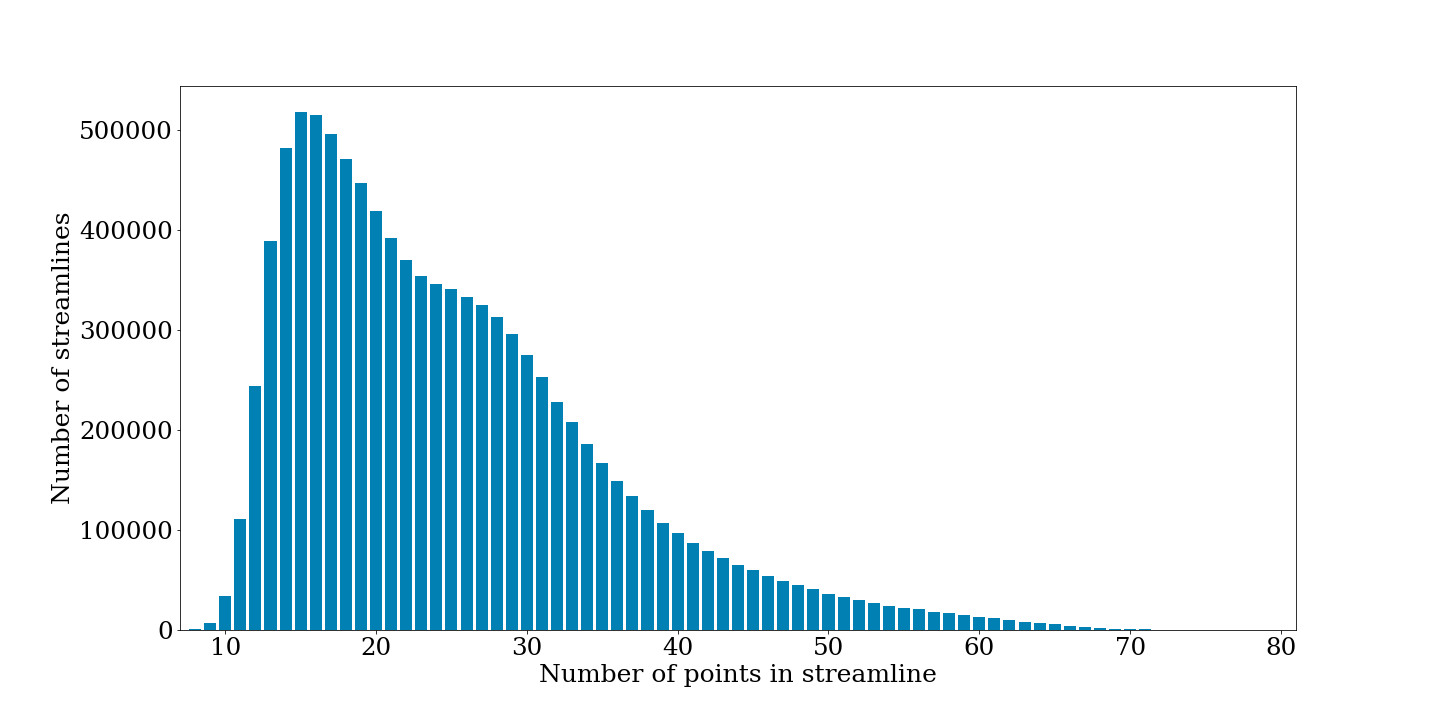}
\includegraphics[width=0.7\textwidth]{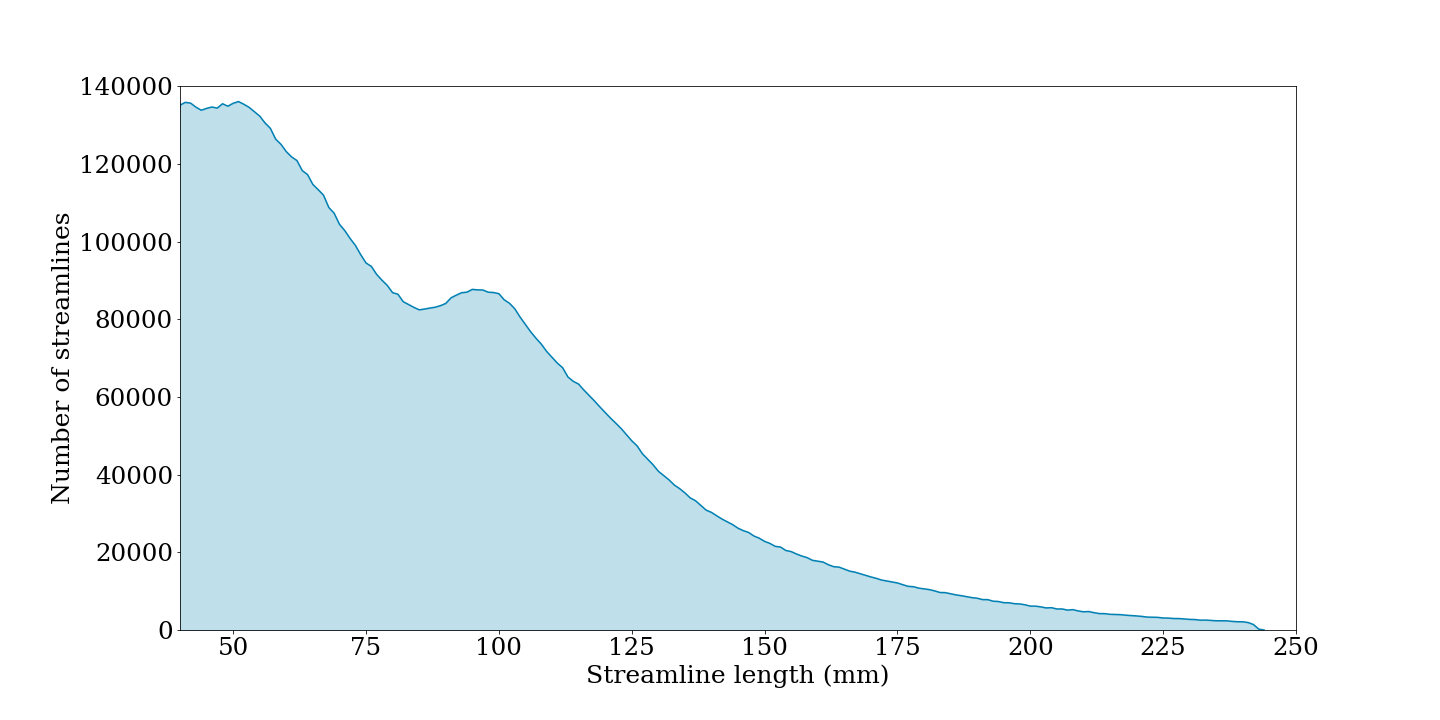}

\includegraphics[width=0.7\textwidth]{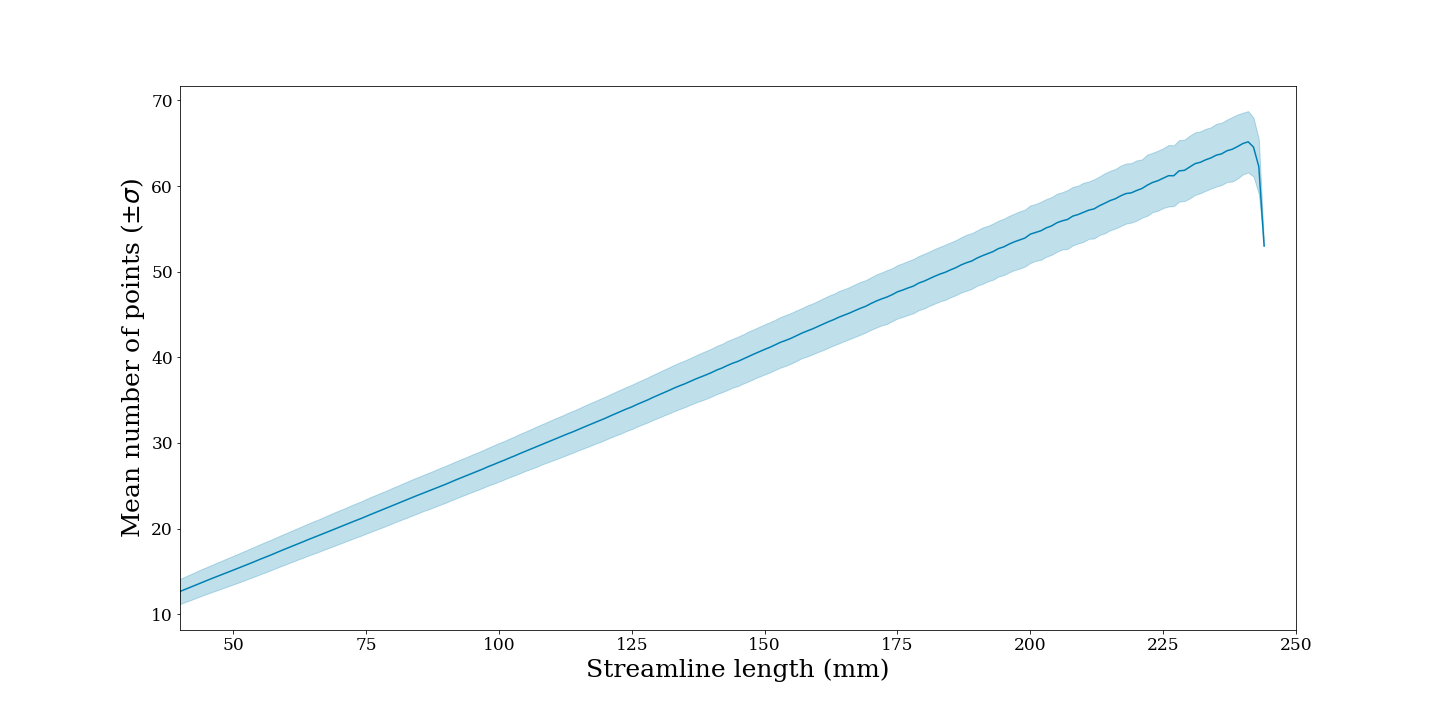}
\caption{Distribution of the number of points and length of streamlines in an exemplary tractogram of ten million streamlines. Top: histogram of the number of points per streamline. Middle: histogram of streamline length in mm. Bottom: correlation plot between streamline length and the mean number of sampling points with standard deviation (light blue). 
}
\label{fig:streamlinelengthsvscoords}
\end{figure}

The ten million streamlines in each HCP tractogram covered the entire white matter volume. An exemplary depiction of the distribution of streamline length, the number of sampling points, and their correlation for one subject can be seen in Fig.~ \ref{fig:streamlinelengthsvscoords}. This figure also shows that the streamline length and the number of sampling points are still highly correlated after compression. The tractograms were generated with MRtrix3 \cite{Tournier:2019} with the exception of the compression step, for which the Python library Dipy \cite{Dipy:2014} was used.

We further studied the phantom data from the DiSCo Challenge \cite{DiSCo:2021}, which was used as ground truth to create datasets with known true positive, redundant, and false-positive streamlines. The ground truth of this dataset consists of 12 196 streamlines.
In our first experiment on this data, we used half of the streamlines as true positives, while the other half was distorted to produce realistic implausible streamlines by applying random rotation in 3D. The rotations were done by randomly selecting Euler angles in the range of 45--315 degrees to get streamlines not too close to the ground truth. We filled the tractogram with false positives to a total of 89 570 streamlines in order to achieve a similar streamline density compared to the 10 million used in the HCP case. 
In a second experiment, half of the ground-truth streamlines were used as true positives, while the second half was systematically copied to produce redundant streamlines. Notice that all streamlines are plausible in this experiment. The second half of the data was split evenly into five further groups: the first fifth of the streamlines was copied once, such that each of them would appear twice in the tractogram. The other groups were copied twice, four, nine, and 48 times (filler data), respectively. This resulted in a tractogram with around 89 000 streamlines, including groups of streamlines appearing with very different frequencies.

\subsection{Randomized SIFT (rSIFT)}

\begin{figure}
    \centering
    \includegraphics[width=\textwidth]{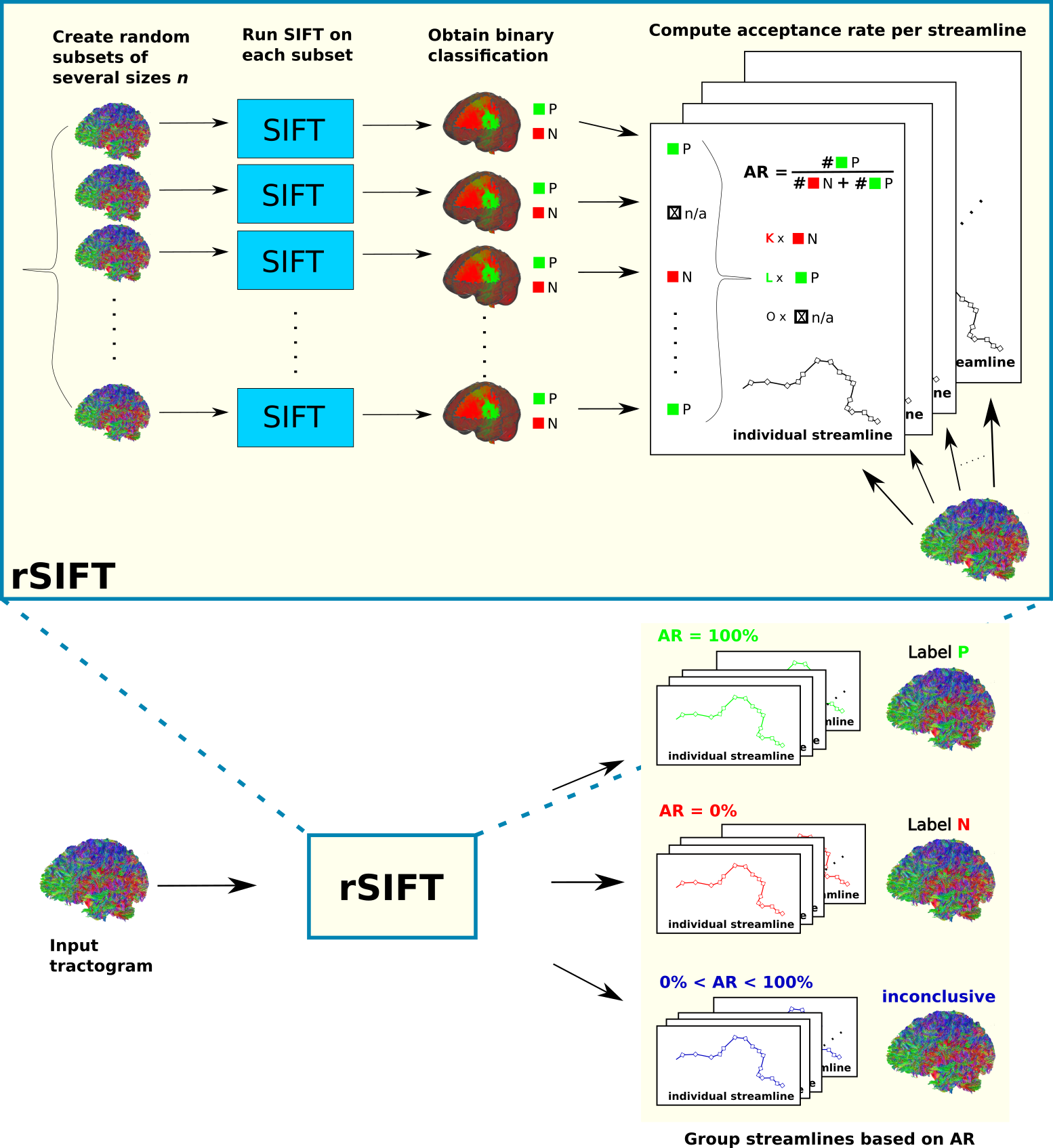}
    \caption{Pipeline of rSIFT. Top: SIFT is run on different random subsets of the original tractogram. The numbers of positive and negative votes are used to estimate the acceptance rate (AR) of every streamline. Bottom: streamlines with AR=100\% and 0\% are used to compute a pseudo ground truth of streamlines. The remaining streamlines are inconclusive.}
    \label{fig:graphical_abstract}
\end{figure}

Fig.~\ref{fig:graphical_abstract} summarizes the method, and Algorithm~\ref{fig:algo} shows the pseudocode of rSIFT. The following subsections detail the different components of the method. Instead of running SIFT once on the complete tractogram, we applied it several times to random subsets of streamlines. The number of times a streamline was filtered out or kept over the different runs of SIFT was used to define its \textit{acceptance rate} (AR) per subset size. We refer to each evaluation of a streamline through SIFT as a \textit{vote}. That means a streamline that was kept in a tractogram after filtering receives a positive vote and a streamline that was removed receives a negative vote.

\begin{algorithm}
\caption{Pseudocode outlining the procedure of rSIFT. For each chosen subset size $n$, $k$ random subsets of the tractogram are extracted and filtered with SIFT to receive the indices of the accepted and rejected streamlines $subset_i^P$ and $subset_i^N$. These are used to update the number of votes.}

\KwIn{
\begin{tabular}{ll}
$T = \{t_1, t_2, \ldots, t_{M}\}$ : tractogram,\\
$SS = \{s_1, s_2, \ldots, s_N\}$ : subset sizes
\end{tabular}}

\KwOut{
\begin{tabular}{ll}
$P = \{p_1, p_2, \ldots, p_M\}$: plausibility votes,\\
$N = \{n_1, n_2, \ldots, n_M\}$: implausibility votes
\end{tabular}}

\BlankLine
\BlankLine
Initialize $P$ and $N$ with zeros\\
\ForAll{$n \in SS$}{
Initialize $P_n$ and $N_n$ with zeros\\
$k \gets \tau M/n$\\
\For{$i \in \{1, \ldots, k\}$}{ 
$subset_i \gets \{r_1, \ldots,r_n\} \subseteq T$, where $ r_{1, \ldots, n}$ are randomly selected from $T$\\
$[subset_{i}^P, subset_{i}^N] \gets SIFT(subset_i)$\\
$P_n(subset_i^P) \gets P_n(subset_i^P)+1$\\
$N_n(subset_i^N) \gets N_n(subset_i^N)+1$
}
$P= P+P_n$\\
$N= N+N_n$\\

}
\label{fig:algo}

\end{algorithm}

In order to examine the influence of not only the composition, but also the subset size on the results of the SIFT algorithm, we employ different subset sizes in the rSIFT procedure. The streamline subset sizes were selected to cover a wide range but simultaneously be able to produce meaningful results within reasonable computation time.

In pretests, out of the ten million streamlines per HCP tractogram, only around 2--2.5\% remained after filtering. The fact that the algorithm terminated through convergence indicates that the remaining set of streamlines was too small to guarantee a stable model and valid removal of streamlines. Any results for streamline sets smaller than that might not be meaningful. Thus, the smallest subset size was chosen to be 2.5\% of the original tractogram size, i.e., 250 000 streamlines for the HCP data. We applied a similar procedure to determine a minimum subset size for the DisCo data, which turned out to be around 16 000 in the tractogram with ground truth and false-positive data and around 13 000 in the tractogram with ground truth and redundant data.

Between the largest (the complete tractogram) and smallest subsets, subset sizes were systematically chosen, usually by halving the subset size. Thus, the probed sizes were $SS = \{1 \times10^7, 5\times10^6, 2.5\times10^6, 1.25\times10^6, 6.25\times10^5, 5\times10^5, 2.5\times10^5\}$ for the HCP data.

For the first DiSCo data experiment (with implausible streamlines), we used $SS = \{$89 570, 44\,785, 22 392, 16 000$\}$ and for the second experiment (with redundant streamlines), $SS = \{$88 996, 44 498, 22249, 13 000$\}$.

Subsequently, rSIFT was run on all of the pre-defined subset sizes $n \in SS$, with a number of repetitions $k$ defined for each $n$. $k$, was adapted to $n$, such that $n \times k$ remained constant. In particular, $k=\tau M/n$, with $M$ being the total number of streamlines in the tractogram and $\tau$ being a parameter that we set to five in the experiments. This way, we aim to obtain enough votes for each streamline to compute robust statistics (e.g., $k=5$ and $k=20$ for $n=1\times10^7$ and $n=2.5\times10^6$, respectively, in the experiments with HCP data).

Notice that, since each streamline subset is defined from the complete tractogram, the number of occurrences across all subsets varies for all streamlines. That means that the number of received votes is expected to differ between streamlines and could, for some, even be zero.

After completing the filtering procedure, the total numbers of positive and negative votes of a streamline $s$ for subset size $n$, denoted as $P_n(s)$ and $N_n(s)$, respectively, were used to compute its acceptance rate $AR_n$ as:
\begin{equation}
AR_n(s) = \frac{P_n(s)}{P_n(s) + N_n(s)}.    
\end{equation}
In addition to this, $AR(s)$ (without $n$) refers to the acceptance rate of a streamline compiling the votes received over \textit{all} subset sizes.
For the analysis of the method, we analyzed the distribution of $AR_n$ over different choices of $n$.

As mentioned, finding an anatomically reliable ground truth is an ongoing challenge in tractogram filtering. Having this in mind, our focus here is limited to extending the notion of streamline acceptance that is possible with SIFT. Streamlines with AR = 100\% and AR = 0\% are likely to be plausible and implausible, respectively, since all runs of SIFT are consistent for these streamlines regardless of the tractogram configuration they are in. That means that these streamlines can be used to generate a \textit{pseudo} ground truth of a plausible and implausible class of streamlines in the sense of SIFT. The plausibility of the remaining streamlines with less consistent results is harder to assess solely based on their AR. Thus, we group them with the label \textit{inconclusive}. Notice that inconclusive streamlines may consist of a mix of plausible and implausible streamlines that SIFT cannot consistently detect or that SIFT has not evaluated in all subset sizes. In summary, we separate the streamlines in a tractogram into three classes: 
\begin{enumerate}
    \item \textit{plausible} where $AR = 100\%$ 
    \item \textit{implausible} where $AR = 0\%$
    \item \textit{inconclusive} where $0\% < AR < 100\%$.
\end{enumerate}
We use the distinction of plausible and implausible streamlines as pseudo ground truth for training the classifiers described in the next section.

\subsection{Neural network-based streamline classifier}

We designed rSIFT with the goal of providing useful information about the plausibility of individual streamlines. In order to gain insight into the properties of the streamline grouping based on plausible, implausible, and inconclusive labels, we analyzed the performance of a neural network classifier trained on these labels in different scenarios. In doing so, we simultaneously show the feasibility of training a neural network model that mimics the characteristics of rSIFT, and thus provide a computationally efficient alternative implementation to the originally proposed method.

In our experiments, we investigated the composition of the three rSIFT labels by means of their separation based on neural network classifiers. First, we trained models for the pair-wise separation of labels, i.e., plausible (positive) vs. implausible (negative), plausible vs. inconclusive, and implausible vs. inconclusive streamlines. Secondly, we trained a multi-class model to distinguish between all three rSIFT labels simultaneously.

The training of all classifiers was performed using a 5-fold cross-validation (CV) approach on the pooled set of streamlines from two HCP subjects. We report the average performance of all CV models on the respective validation set. For testing, we chose the best performing model in terms of balanced sensitivity and specificity in order to minimize bias. This model was used to evaluate streamlines from four unseen HCP subjects, which we refer to as test data.

In order to make the classifier applicable to data from multiple subjects and to provide a basic image registration, streamline coordinates were normalized to a range of -1 and 1 in each dimension using the minimum and maximum streamline coordinates per subject.
Since the input streamlines comprised a varying number of points, they were resampled to the same number of points in order to be used as input for the network. To minimize the impact of this resampling on as many streamlines as possible, the median number of streamline points across training subjects, which was determined to be 22 points, was chosen as the resampling target. Linear interpolation was applied to approximate the original streamline geometry.

\begin{figure}
\centering
\begin{tabular}{ccc}
\includegraphics[scale=0.45]{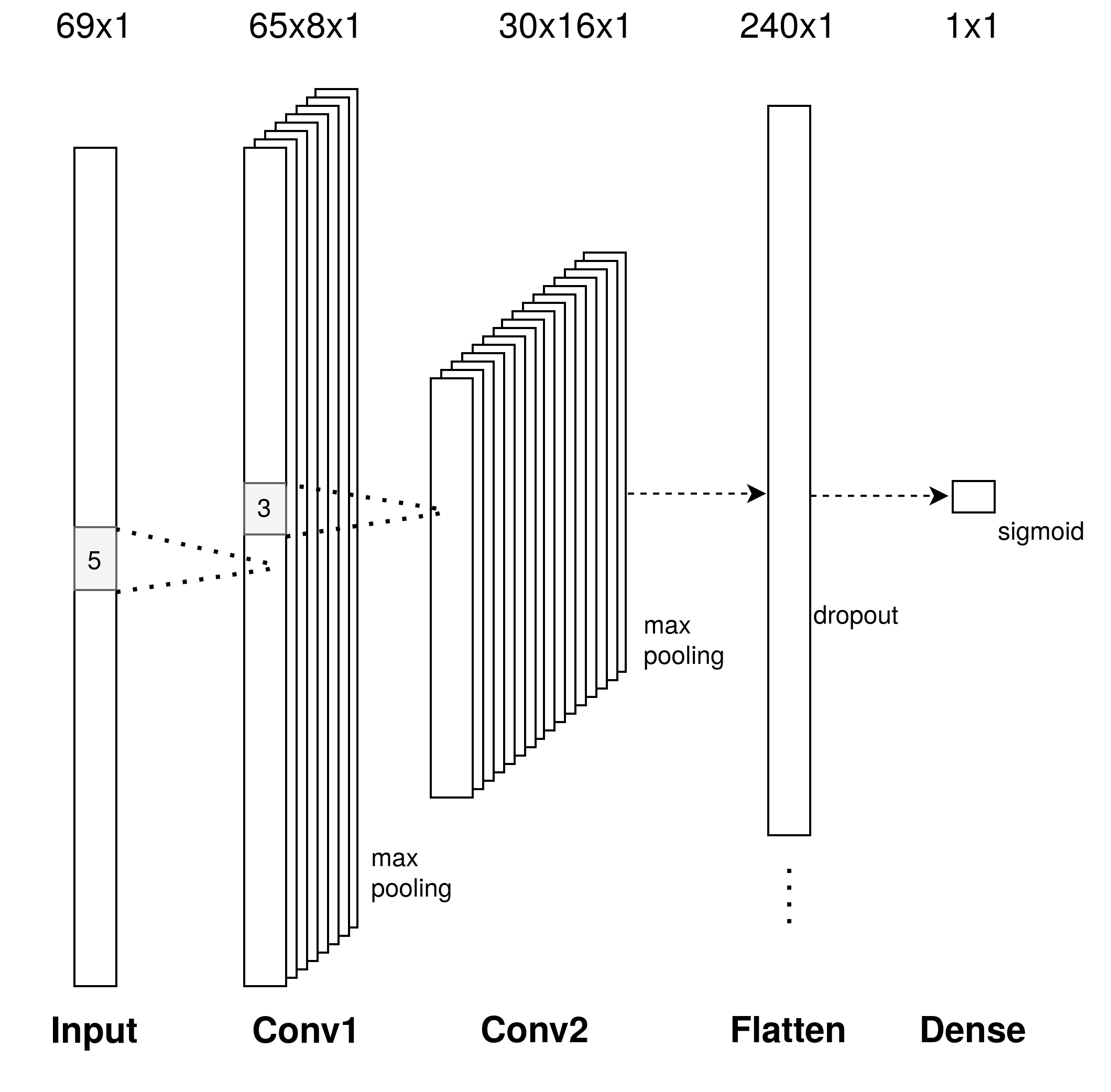} & \includegraphics[scale=0.45]{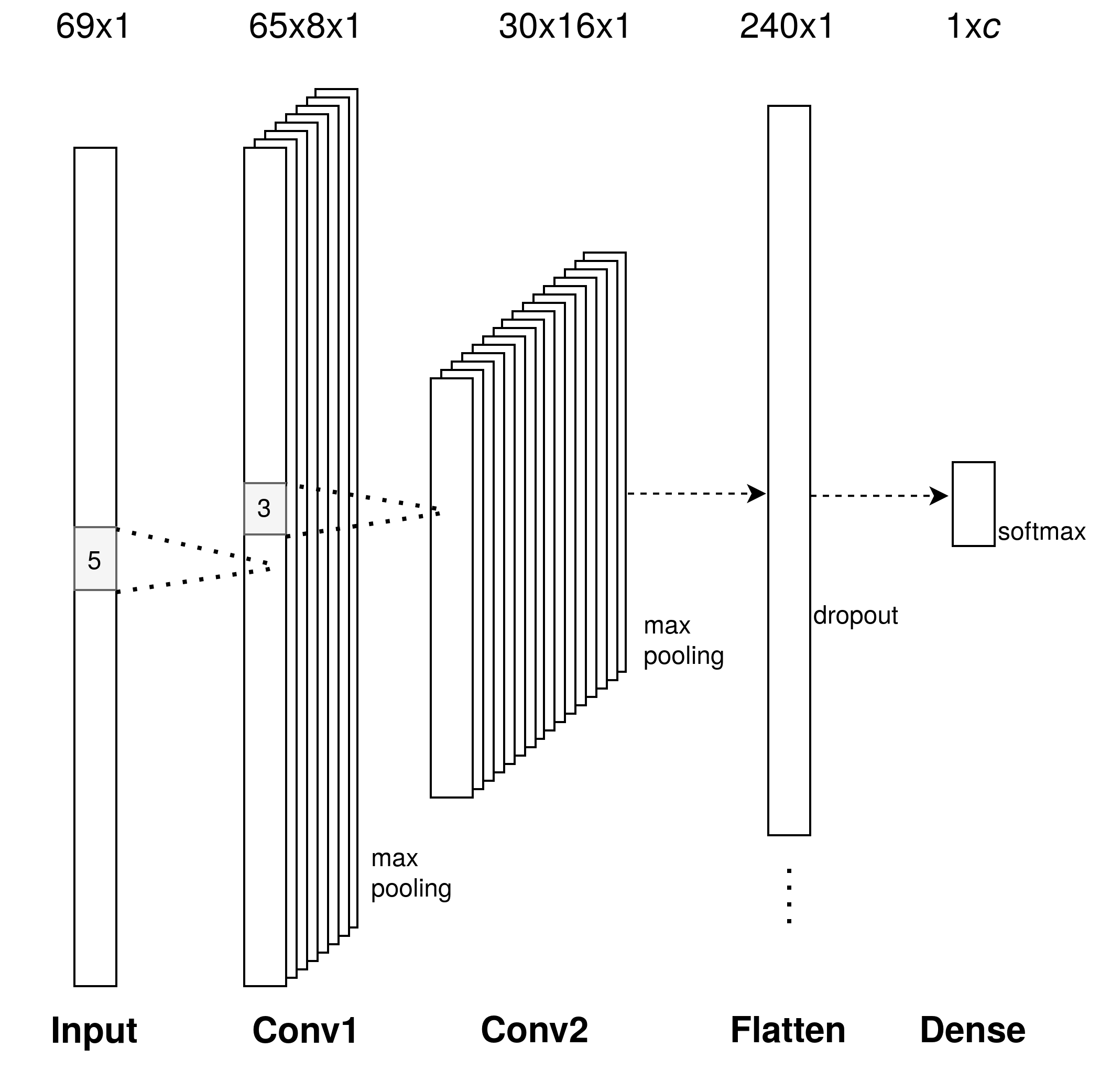}\\
\end{tabular}
\caption{Architecture of the classifiers. Left: binary classifier. Right: multi-class classifier. The only difference is in the last layer. The binary classifier uses one output neuron with sigmoid activation, while the multi-class classifier uses multiple output neurons with softmax activation.}
\label{fig:classifiers}
\end{figure}

Each classifier was composed of two 1-D convolution layers (kernel sizes 5 and 3), with ReLU activation and max-pooling (pool size 2) applied after each of them. The convolutional layers were followed by a dense layer with a dropout chance of 0.5 and connected to either one or multiple output neurons (depending on the classifier type being binary or multi-class). For the binary type, the last layer used a sigmoid activation function, while the multi-class classifier used softmax. Therefore, both classifier architectures were identical except for the number of output neurons (one vs. multiple) and the corresponding activation functions. An illustration of the structure of the classifiers is shown in Fig.~\ref{fig:classifiers}. The input to the network consisted of a one-dimensional array of the coordinates of one streamline reordered in an interleaved manner, such that the x-, y-, and z-coordinates of the same point were subsequently following each other (i.e., $x_1, y_1, z_1, x_2, y_2, z_2, ...$).

Since the data labels were determined based on a filtering method that rejects the vast majority of streamlines, a large disproportion in the number of training samples available for each class was expected. Thus, a balanced data generator was used to oversample data of plausible streamlines for network training and testing. The epoch size was defined by the larger of the classes such that each training sample was seen at least once per epoch. The smaller set of samples was shuffled and re-used whenever it was exhausted during an epoch. 
Each classifier was trained towards maximizing accuracy using the Adam optimizer \cite{Kingma:2017} with default settings in TensorFlow 2.8.0 \cite{Kingma:2017}.
The training was done in batches of 50 samples each (except for the categorical classifier with three classes, which used a batch size of 60 such that each batch could contain an equal number of samples from each class). Five training epochs were determined to be sufficient since accuracy and loss saturated quickly and showed almost no improvement after two epochs.

\section{Results}

\subsection{Randomized SIFT on HCP data}

Fig.~\ref{fig:table_1} and \ref{fig:votes graph} show the distribution of streamlines for the different subset sizes and AR ranges on the six HCP subjects. As shown, most of the streamlines received either $AR_n = 0\%$ or $AR_n = 100\%$ after rSIFT for all $n$.
Notice that the fractions of those streamlines with $0\% < AR_n < 100\%$ increased when the subsets became smaller. Moreover, using smaller subset sizes led to a substantial increase in streamlines with $AR_n = 100\%$ even after repeated evaluation: The number of streamlines with exclusively positive votes rose from an average of 2.63\% to 14.86\%, comparing the largest and the smallest subset sizes. Conversely, the streamlines with $AR_n = 0\%$ were reduced from 96.64\% to 53.67\%. As shown in the last column of Fig.~\ref{fig:table_1}, the number of plausible and implausible streamlines for the whole dataset are around 1.7\% and 53.7\%, which means that approximately 44.6\% of the streamlines can be considered as inconclusive. Interestingly, SIFT yielded inconclusive results for on average 0.7\% of the streamlines when the complete tractogram was used (i.e., subset size of ten million), in four of the HCP subjects.

As mentioned, the rSIFT procedure does not guarantee having the same amount of votes for each streamline. Thus, we assessed the distribution of ARs for streamlines with exactly five votes, as shown in Fig.~\ref{fig:table_votesfive}. Although the values are slightly different in this case, they follow a very similar trend. Due to the entirely random choice of streamlines for each round of the experiment, we found that the number of streamlines that were not included in any of the subsets in each subset size was negligible. They were not included in Fig.~\ref{fig:table_1}.
Every streamline received on average 35 votes over all experiments, with an average of five votes per subset size.

\input{submission_random_res.tex}

\begin{figure}

\centering
\small
\includegraphics[width=\textwidth]{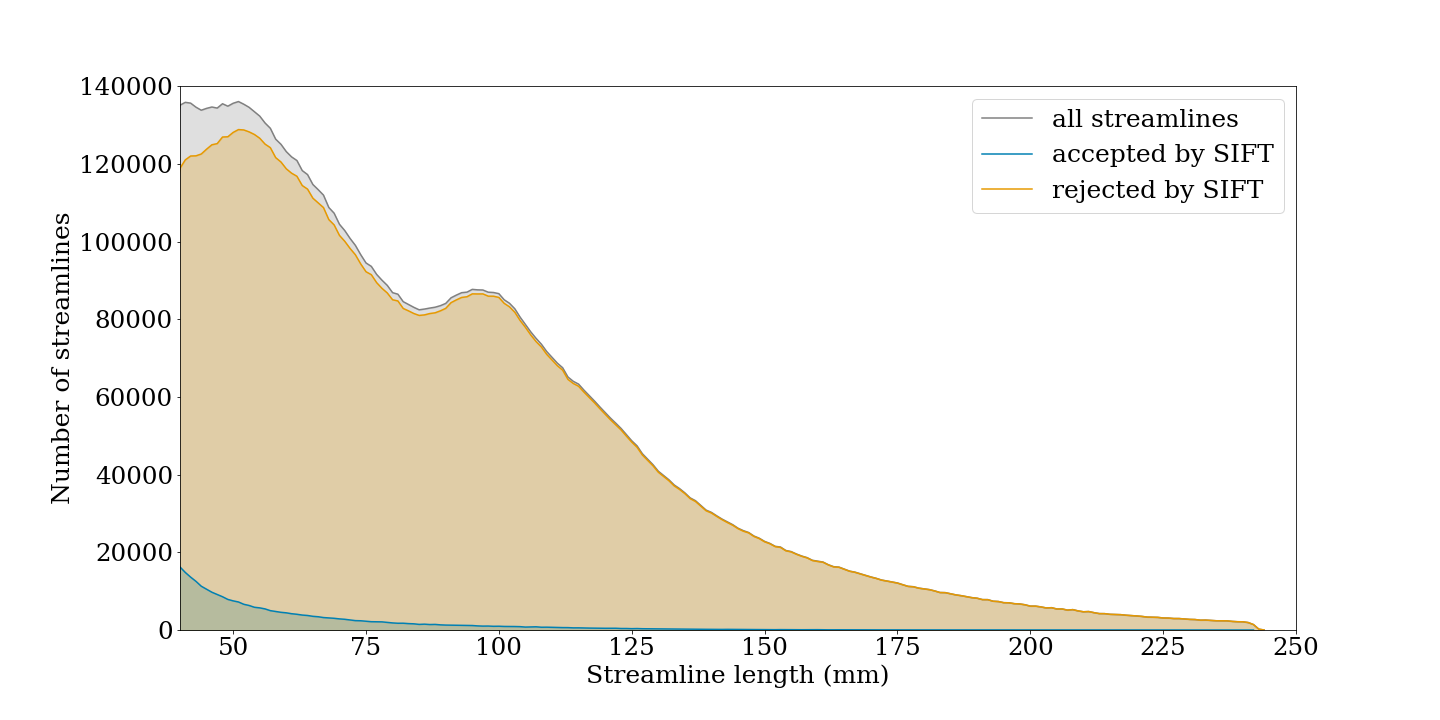}
\caption{Distribution of streamlines accepted and rejected by SIFT with respect to their length for one exemplary subject with ten million streamlines.}
\label{fig:lenghtssift}
\end{figure}

\begin{figure}

\centering
\small
\includegraphics[width=\textwidth]{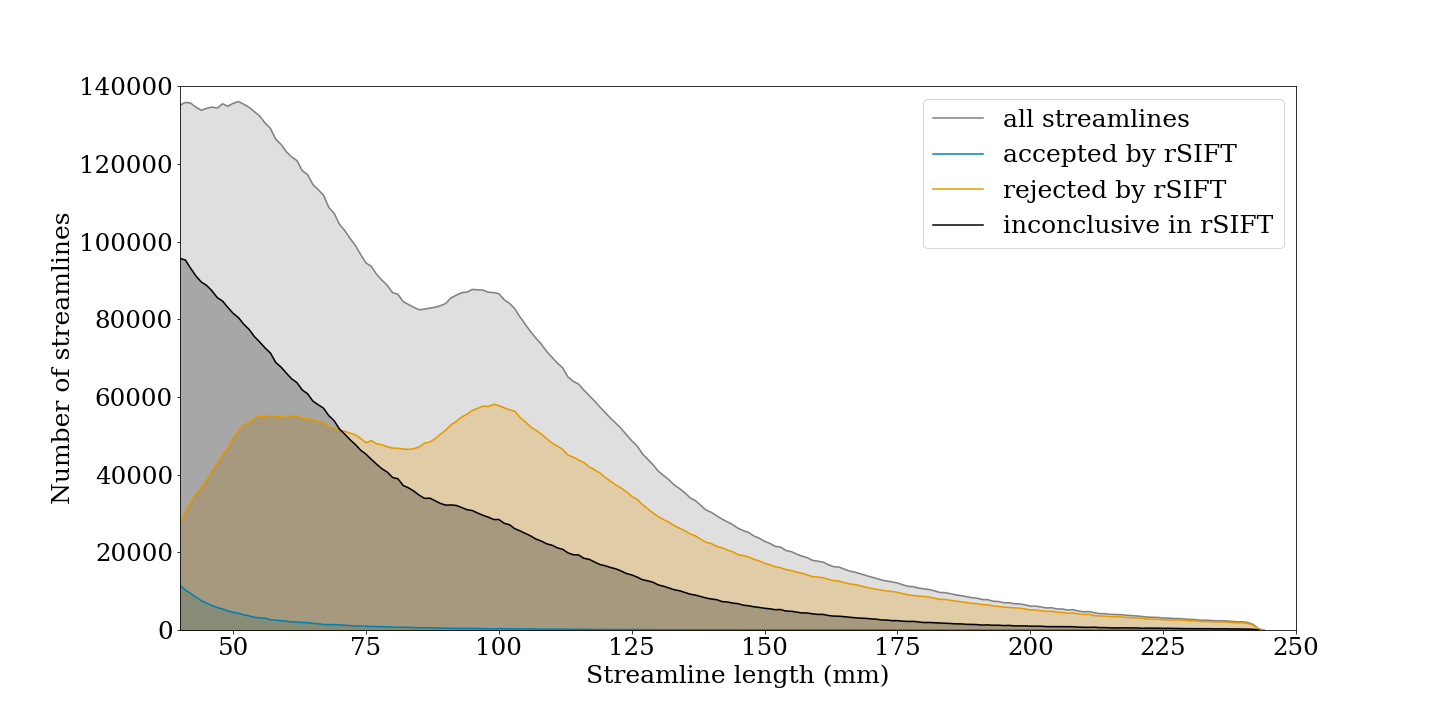}
\caption{Distribution of streamlines labeled plausible (accepted), implausible (rejected), and inconclusive by rSIFT with respect to their length for one exemplary subject with ten million streamlines.} 
\label{fig:lengthsSIFTvsrand}
\end{figure}

We were further interested in assessing the influence of the streamline length on the filtering. As shown in Fig.~\ref{fig:lenghtssift}, 
SIFT strongly influenced the distribution of streamline lengths. In fact, streamlines deemed plausible were among the shortest and barely exceeded the length of 125mm. The same pattern is shown for rSIFT in Fig.~\ref{fig:lengthsSIFTvsrand}. Notice that the new class of inconclusive streamlines covers the whole range of lengths.

\subsection{Randomized SIFT on the DiSCo data}

\begin{figure}
    \centering
    \includegraphics[width=0.8\textwidth,clip,trim=5.5cm 0cm 0cm 0cm]{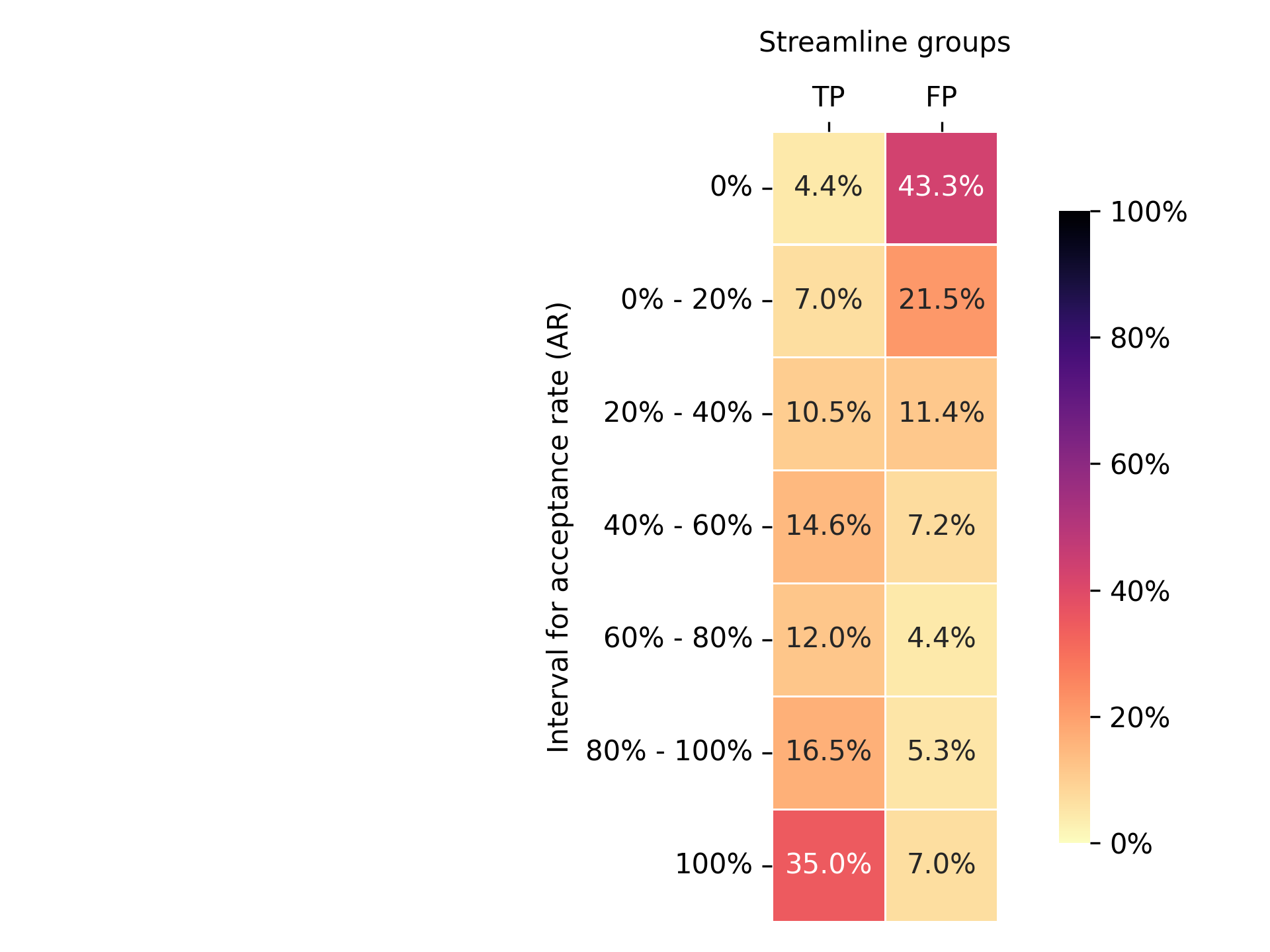}
    \caption{Distribution of acceptance rate (AR) for
true positive (TP) and false positive (FP) streamlines in the DiSCo dataset after running rSIFT with $\tau=5$ and four different subset sizes. The range 80--100\% excludes $AR = 100\%$. Note that the values in the columns TP and FP sum to 100\%, respectively.}
    \label{fig:disco_tpfp}
\end{figure}

Fig.~\ref{fig:disco_tpfp} summarizes the ARs of rSIFT on the DiSCo data with false positives, considering all subset sizes. Although SIFT filters out false positives more often, it also filters out true positives. As shown, selecting a single threshold of AR for distinguishing between true and false positives is challenging since it involves a compromise between filtering out as few plausible streamlines as possible at the cost of accepting more implausible ones or filtering out more plausible ones to minimize the false positives. For example, using a threshold of 20\% will lead to a rejection of 64.8\% of false positives, at the cost of losing 11.4\% of true positives in this experiment. In turn, a threshold of 80\% will reject 87.7\% false positives, but, at the same time, it will reject almost half of the true positives (48.5\%). As in the case of the HCP data, analyzing the streamlines independently where SIFT is not consistent might be beneficial. 

\begin{figure}
    \centering
    \includegraphics[width=\textwidth,clip,trim=0cm 0cm 0cm 0cm]{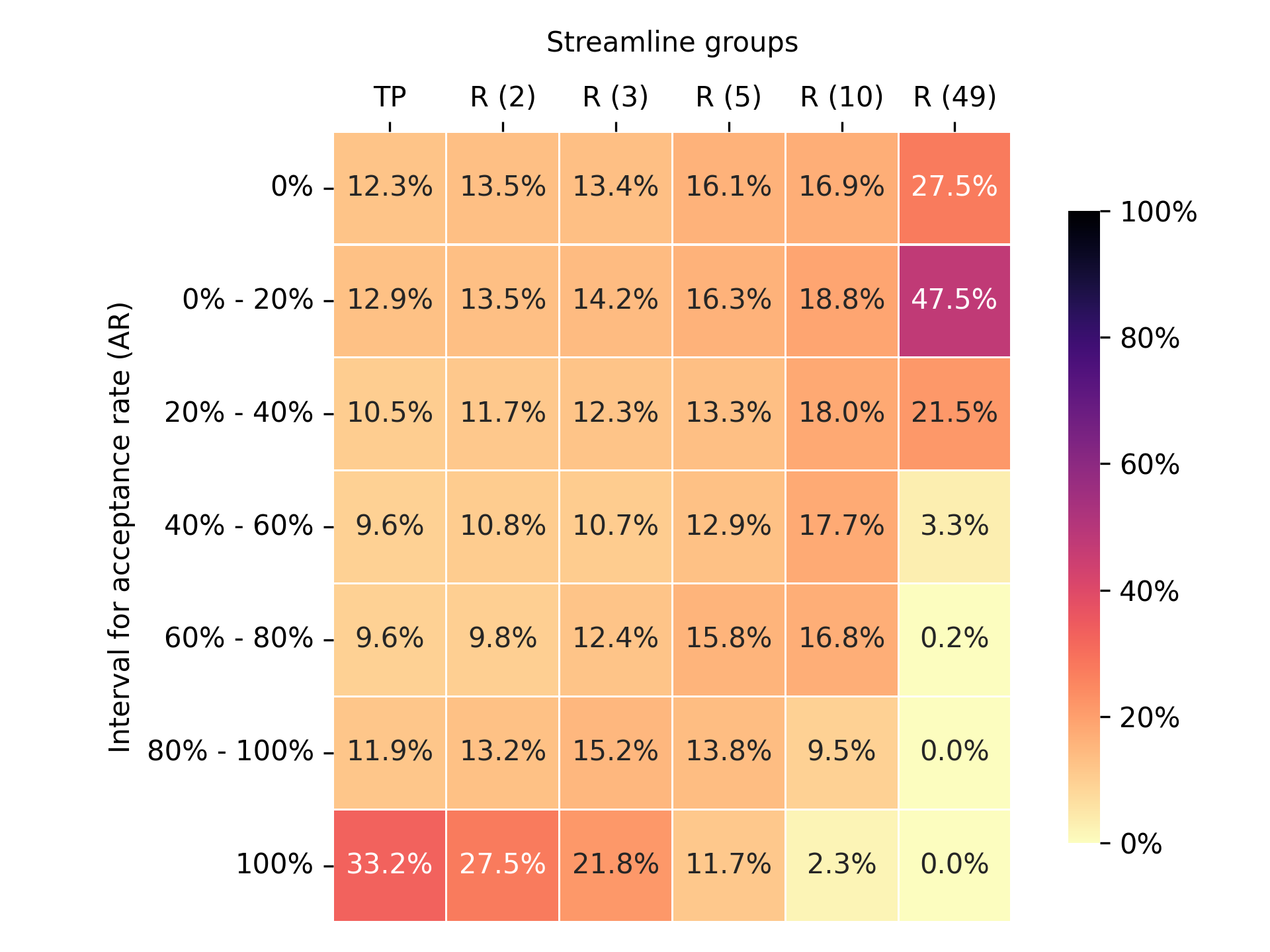}
    \caption{Distribution of acceptance rate (AR) for true positive (TP) and redundant (R) streamlines in the DiSCo dataset
after running rSIFT with $\tau=5$ and four different subset sizes. The number in parentheses indicates the number of replications per streamline, i.e., how often they were found in the tractogram. The range 80--100\% excludes $AR = 100\%$. Each column sums up to 100\%.}
    \label{fig:disco_red}
\end{figure}

Fig.~\ref{fig:disco_red} shows the results of the experiment with DiSCo data with redundant streamlines. Redundant streamlines are filtered out more often when they appear more frequently in the tractogram. For example, the amount of streamlines with $AR = 0\%$ increases from 13.5\% to 27.5\% when their multiplicity grows from 2 to 49, respectively. However, this increment in the amount of rejected streamlines is relatively slow. For example, for $AR = 0\%$, the amount of streamlines only increases by 3.4\% (13.5\% vs. 16.9\%) when increasing the number of redundant copies from 2 to 10. Redundancy has a larger effect at the other end: the number of streamlines with $AR = 100\%$ decreases rapidly with the number of redundant copies.
Comparing Fig. \ref{fig:disco_tpfp} and \ref{fig:disco_red}, it can be seen that SIFT rejects more often true positives when there are no false positives. 
As a comparison, around 53\% of streamlines (6 483 of 12 196) remain in the tractogram when SIFT is used only once on the DiSCo ground truth streamline set.

\subsection{Classification performance}

\begin{figure}
    \centering
    \includegraphics[width=\textwidth,clip,trim=0cm 0cm 0cm 0cm]{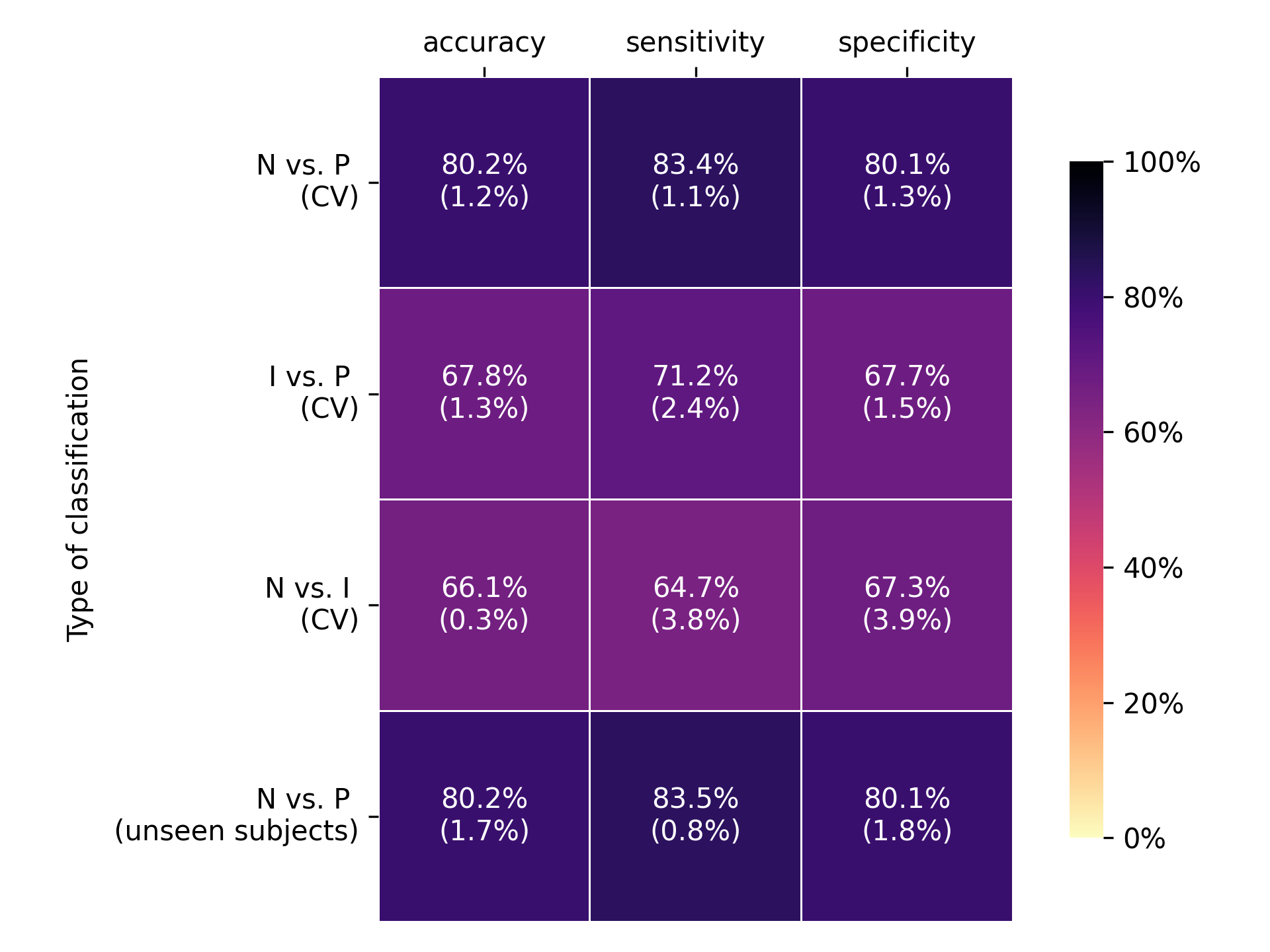}
    \caption{Classifier performance for binary classification. ``N'' denotes ``negative'' (implausible), ``P''  ``positive'' (plausible), and ``I'' inconclusive. The metrics are specified through the mean values as determined by 5-fold cross-validation (CV) in training or tests with unseen subjects during training, with standard deviations in parentheses.}
    \label{fig:classifier}
\end{figure}

An overview of the performance of the three binary classifiers on HCP data can be found in Fig.~\ref{fig:classifier}.

As shown, the binary classifier for plausible and implausible streamlines was around 80\% accurate on the validation folds, with similar accuracy for both classes. This shows that the procedure for balancing the classes was appropriate for this task. 
In turn, the performance of the binary classifiers trained to distinguish inconclusive streamlines from plausible or implausible was much lower and very similar (e.g., accuracy was 67.80\% and 66.09\%, respectively). 

As mentioned, we also evaluated the best-performing CV model for distinguishing negative and positive streamlines on the four unseen subjects. As shown in Fig.~\ref{fig:classifier}, the results of this test were equal to the five-fold CV experiment, with a mean accuracy of 80.2\%.

\begin{figure}
    \centering
    \includegraphics[width=\textwidth,clip,trim=0cm 0cm 0cm 4.5cm]{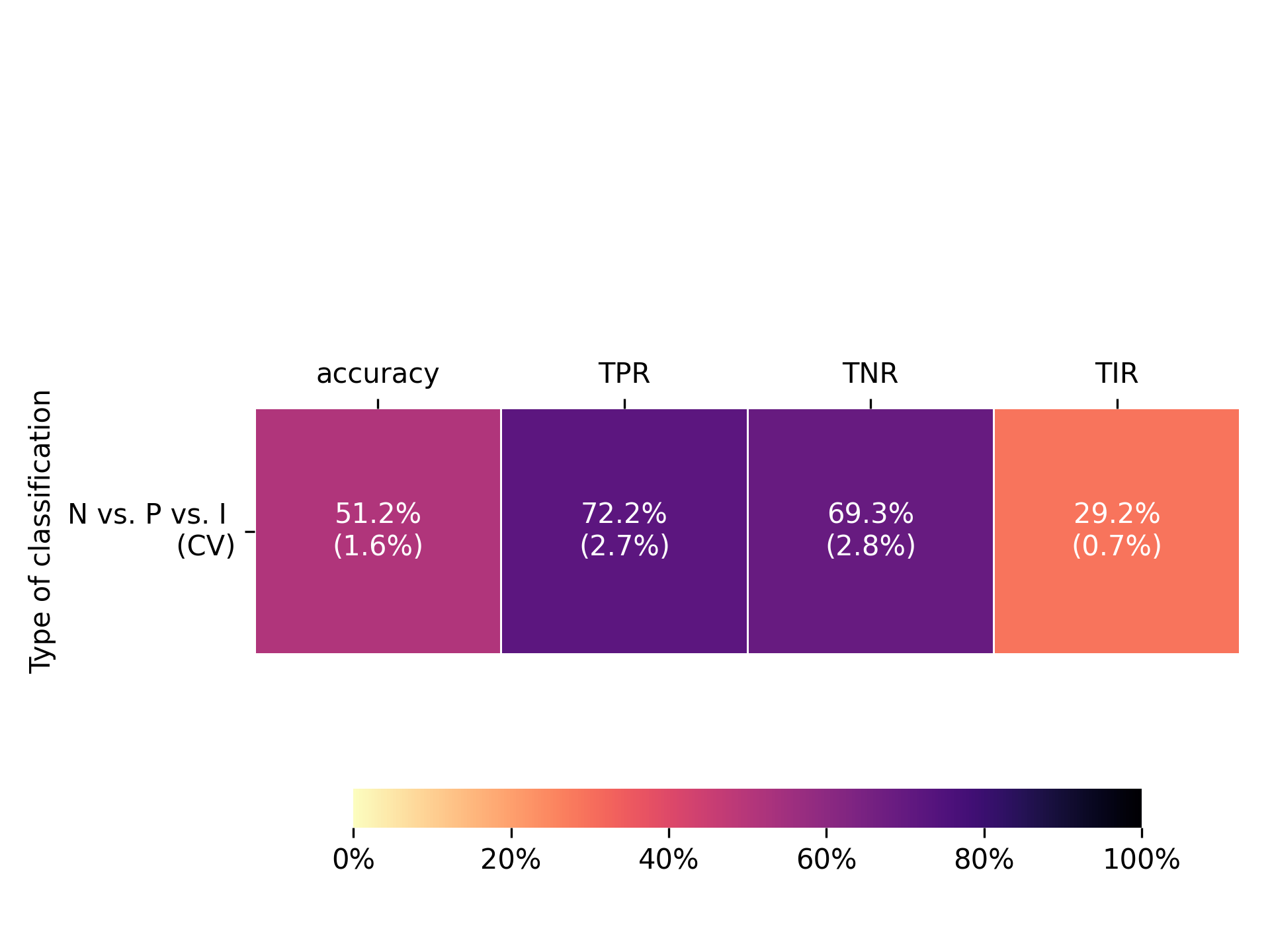}
    \caption{Performance metrics of the multi-class classifier. ``N'' denotes ``negative'' (implausible), ``P''  ``positive'' (plausible) and ``I'' inconclusive. TPR = true positive rate, TNR = true negative rate, TIR = true inconclusive rate.}
    \label{fig:classifiercat}
\end{figure}

\begin{figure}
    \centering
    \includegraphics[width=\textwidth,clip,trim=0cm 0cm 0cm 0cm]{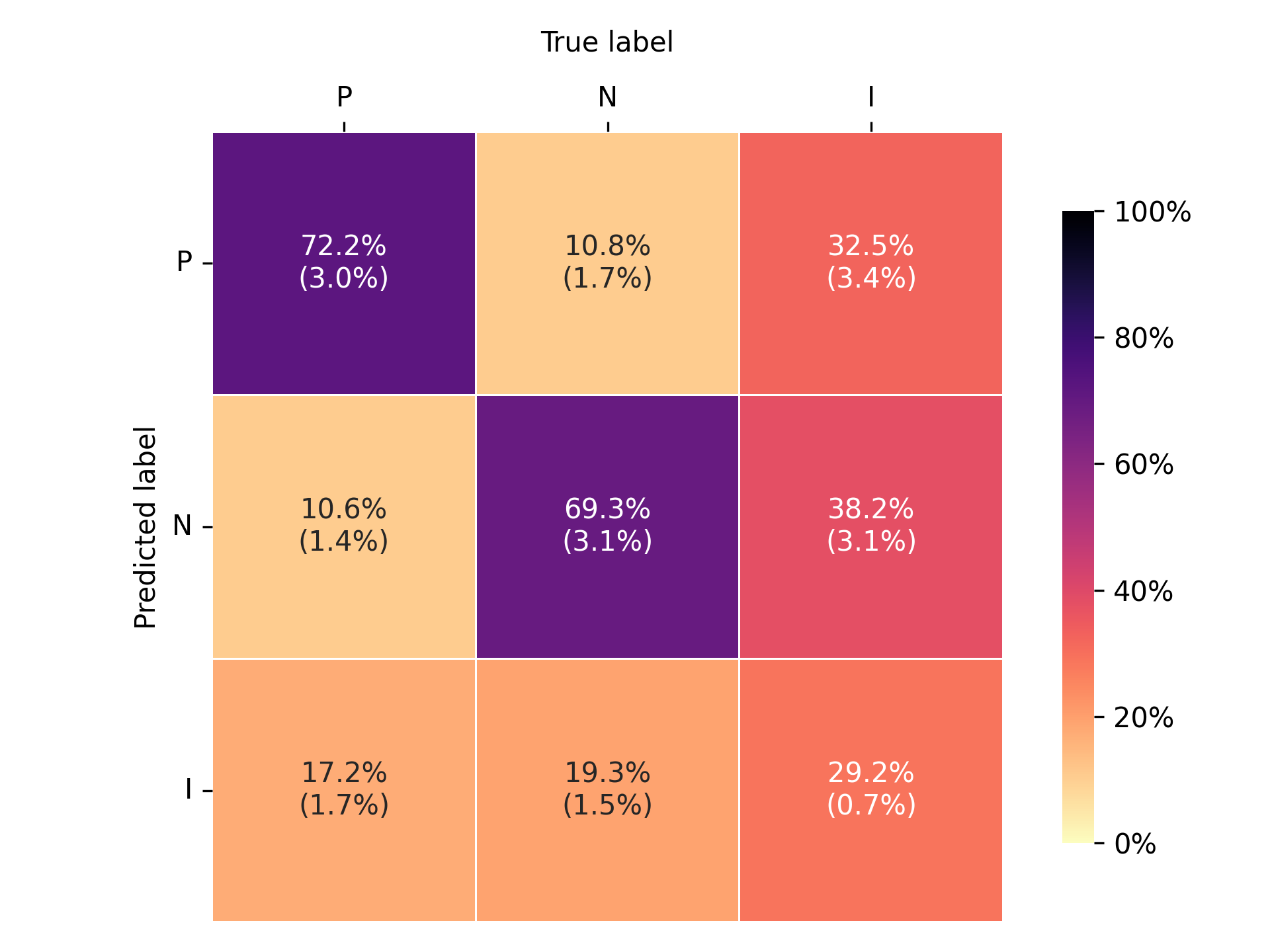}
    \caption{Confusion matrix for the multi-class task involving samples from all three sets from rSIFT. ``N'' denotes ``negative'' (implausible), ``P''  ``positive'' (plausible), and ``I'' inconclusive. The cells show the mean percentage of streamlines belonging to this case in cross-validation, with the standard deviation in parentheses. The columns show the original labels, and the rows show the label given by the classifier. Each column sums up to 100\%.}
    \label{fig:classifiercatconfusion}
\end{figure}

Fig. \ref{fig:classifiercat} and \ref{fig:classifiercatconfusion} show the performance of the multi-class classifier in terms of the confusion matrix and accuracy-based metrics, respectively. This network was trained to recognize the three classes 'P', 'N', and 'I'. While it showed some success on the sets of plausible and implausible streamlines (true positive rate (TPR) and true negative rate (TNR) around 70\% in Fig.~\ref{fig:classifiercat}, column maximum for 'P' and 'N' on the diagonal of confusion matrix in Fig.~\ref{fig:classifiercatconfusion}), its performance remained at chance level for the inconclusive class (true inconclusive rate (TIR) less than $1/3$ for three classes in Fig.~\ref{fig:classifiercat}, similar values in all entries of column 'I' in confusion matrix in Fig.~\ref{fig:classifiercatconfusion}). Therefore, it seems that this classifier is unable to tell the inconclusive streamlines apart from plausible or implausible ones in our pseudo ground truth.

For further analyses, we chose the binary model between plausible and implausible streamlines from the cross-validation fold that yielded the most balanced sensitivity/specificity values (81.26\% and 83.15\%). Raw classifier scores for test samples from the positive class showed a mean of around 0.71 and a median of around 0.76, while the negative test samples received a mean score of around 0.28 and a median of around 0.21. These results suggest that the classifier was relatively confident about the results. However, the mismatch between mean and median hints at the presence of outlier streamlines. Accordingly, classification scores had both a mean and a median of around 0.48 when the classifier was applied to the inconclusive set, which was not part of the training data.

We further split this set into two groups using a threshold of 0.5 on the classification score to get a set of leaning positive and leaning negative streamlines. The mean scores of these two groups were 0.72 and 0.21, which shows that they were slightly more off-center for the negative class.

\begin{figure}
\centering
\small
\includegraphics[width=0.8\textwidth]{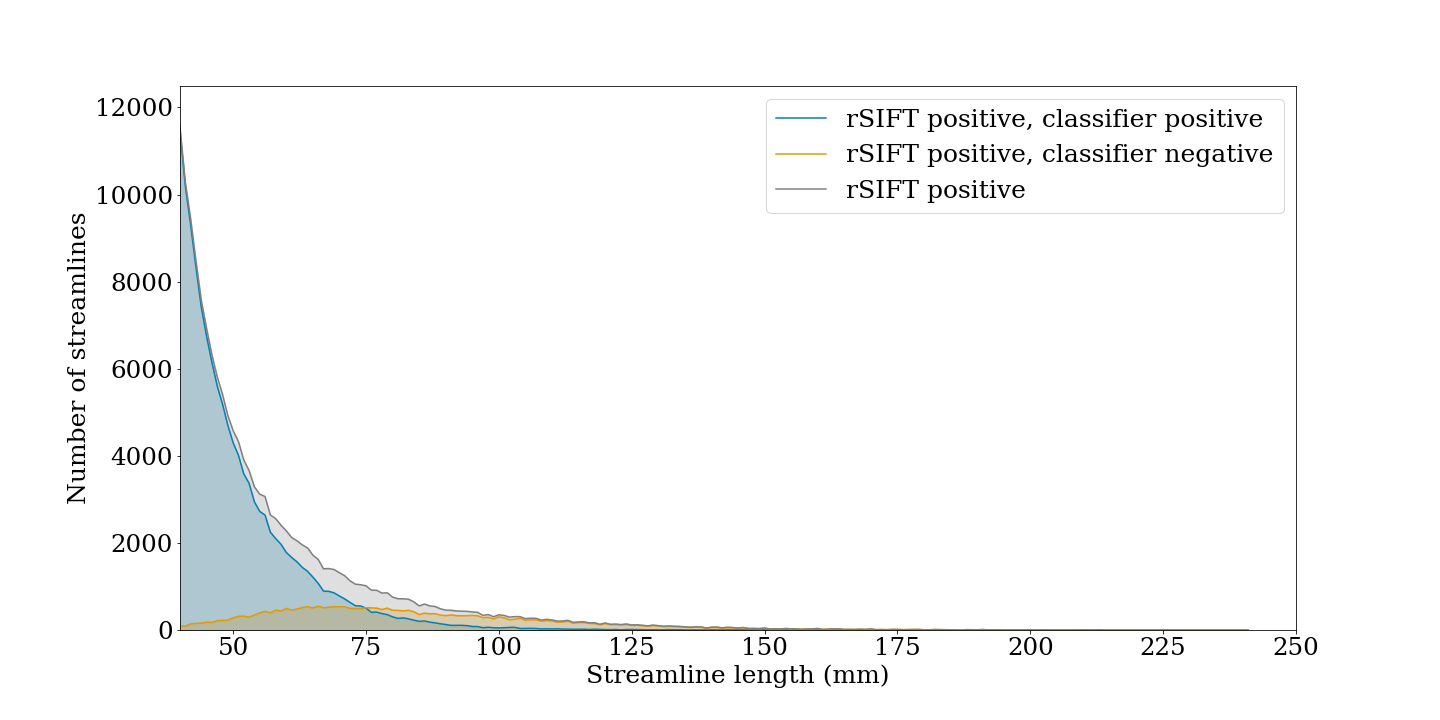}
\caption{Histogram of streamline lengths of positive (P) streamlines, grouped by the streamline labels obtained from the binary positive vs. negative (plausible vs. implausible) classifier. The streamlines were taken from one test subject unseen in training. Longer streamlines are more frequently misclassified (i.e., labeled differently by the classifier than by rSIFT). Negative classifications outnumber positive classifications above a streamline length of around 75mm.}\label{fig:lengthsclassifierp}
\end{figure}

\begin{figure}
\centering
\small
\includegraphics[width=0.8\textwidth]{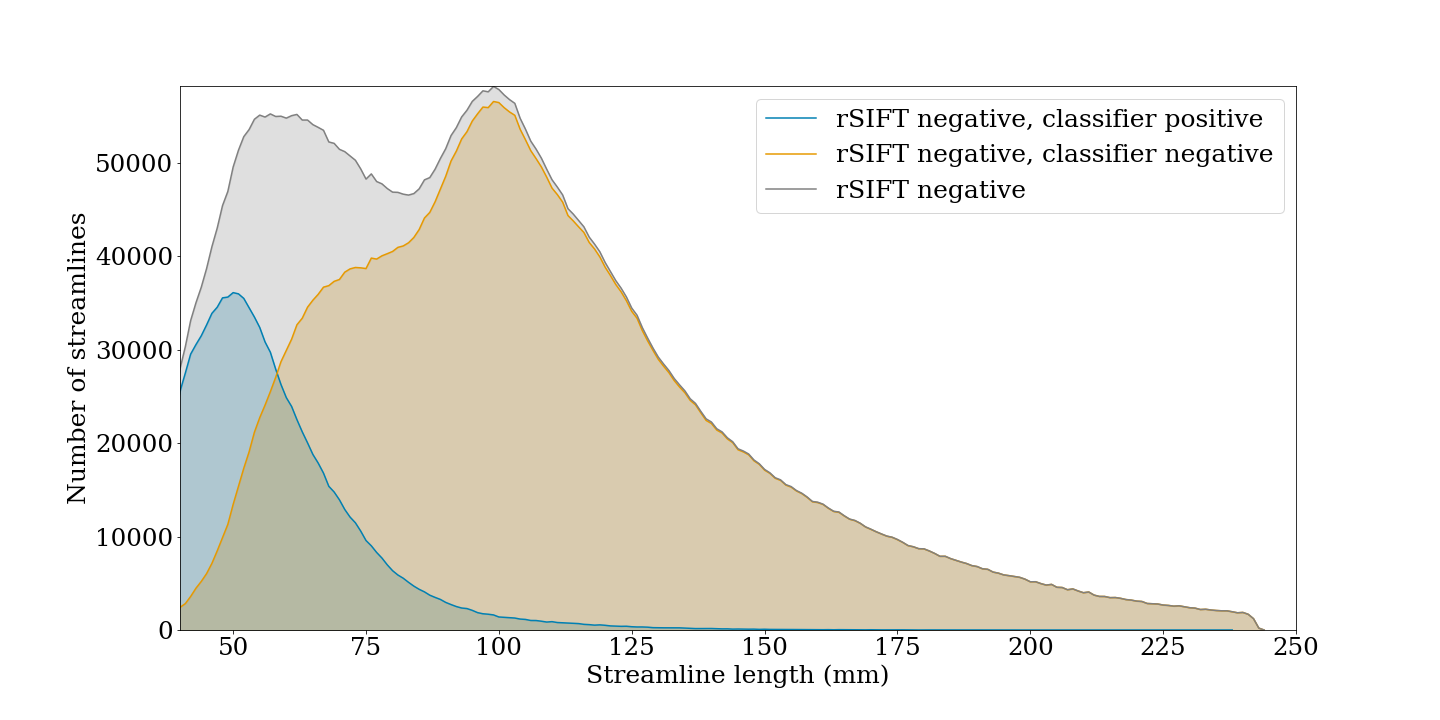}
\caption{Histogram of streamline lengths of negative (N) streamlines, grouped by the streamline labels obtained from the binary positive vs. negative (plausible vs. implausible) classifier. The streamlines were taken from one test subject unseen in training. Shorter streamlines are more frequently misclassified (i.e., labeled differently by the classifier than by rSIFT). Positive classifications outnumber negative classifications below a streamline length of around 60mm.}
\label{fig:lengthsclassifiern}
\end{figure}

\begin{figure}
\centering
\small
\includegraphics[width=0.8\textwidth]{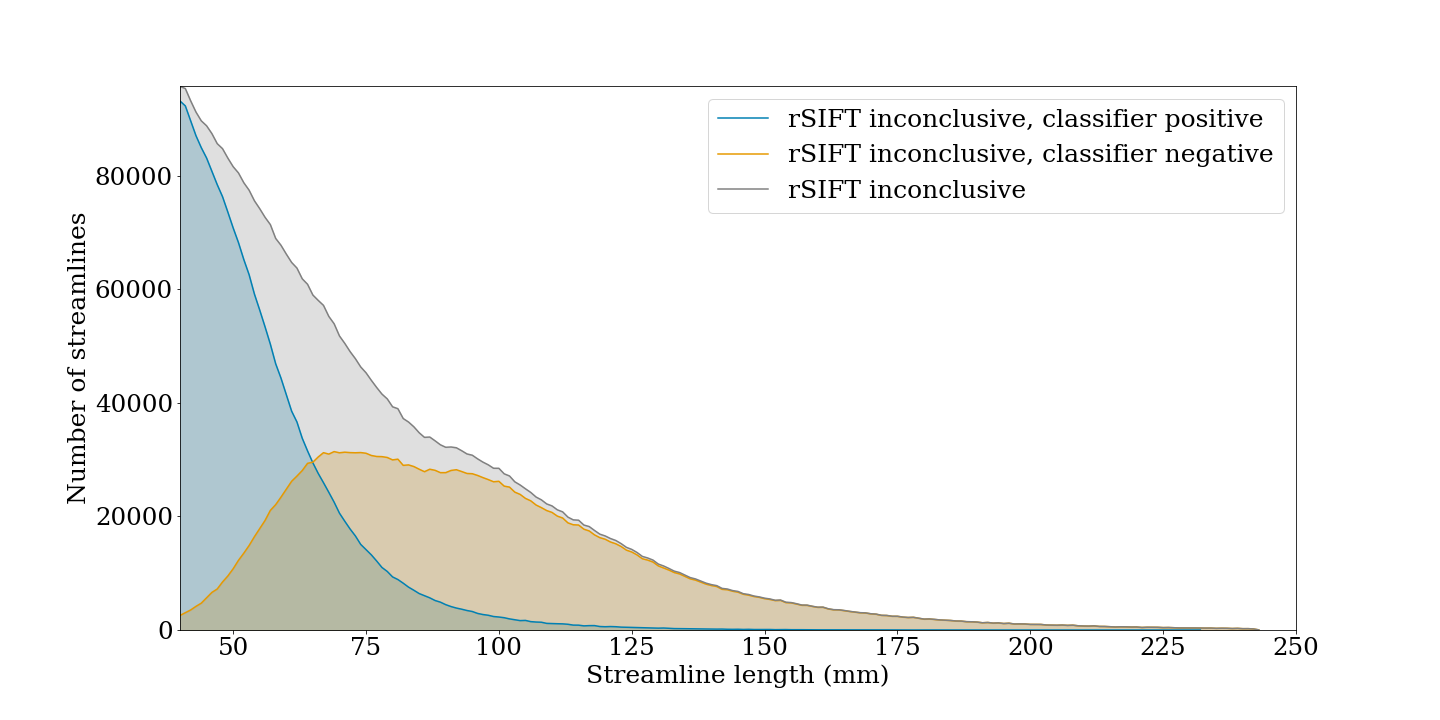}
\caption{Histogram of streamline lengths of inconclusive (I) streamlines, grouped by the streamline labels obtained from the binary positive vs. negative (plausible vs. implausible) classifier. The streamlines were taken from one test subject unseen in training. Shorter streamlines are more frequently classified as positive, and longer streamlines are more frequently as negative, with an equal amount of positive and negative predictions around a streamline length of around 65mm.}
\label{fig:lengthsclassifiero}
\end{figure}

As mentioned before, SIFT tends to filter out long streamlines. Fig.~ \ref{fig:lengthsclassifierp} and \ref{fig:lengthsclassifiern}
show the performance of the binary classifier (P vs. N) with respect to the length of the streamlines. It is apparent that longer streamlines of the positive class were misclassified more frequently (cf. Fig.~ \ref{fig:lengthsclassifierp}), and the same applies to the shorter streamlines of the negative class (cf. Fig.~\ref{fig:lengthsclassifiern}). Applying the P vs. N classifier to the inconclusive set, we found that, also here, shorter streamlines tended more to be classified as plausible and longer ones to be classified as implausible (cf. Fig.~\ref{fig:lengthsclassifiero}). 

\subsubsection{Comparison between classifier scores and rSIFT}

\begin{figure}
\centering
\includegraphics[width=\textwidth]{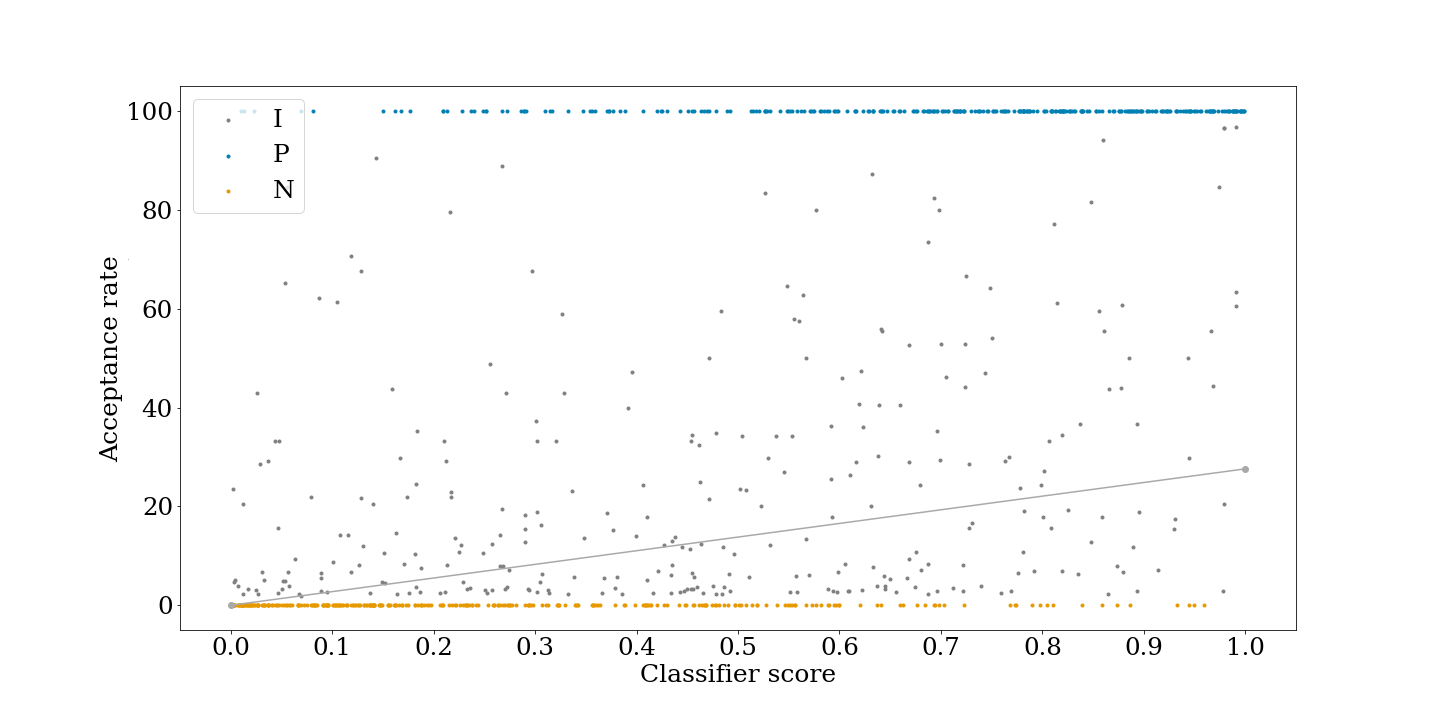}
\caption{Scatter plot relating classifier scores of streamlines and their acceptance rate from rSIFT. The data points (i.e., streamlines) in the plot are sampled in a balanced manner such that all classes (i.e., plausible, implausible, and inconclusive) appear with the same frequency, where blue represents the positive, orange the negative, and gray the inconclusive class. As seen, scores for the positive and negative classes accumulate around the ends of the spectrum. For the inconclusive samples, there is a weak correlation of 0.28 (depicted with the trend line) between scores and AR from rSIFT.}
\label{fig:scatter}
\end{figure}

The correlation coefficient for the relation between classifier scores and the ARs of inconclusive streamlines in rSIFT was moderately positive with a value of 0.28. Fig.~\ref{fig:scatter} depicts scores given by the classifier to each class and how they compare to the vote distribution from rSIFT.

In order to visualize clustering patterns of the three different sets, we used a t-stochastic neighbor embedding (t-SNE) \cite{Maaten:2008}, which enables high-dimensional input vectors to be projected into a 2-dimensional plane. This way, hidden structures in the data as well as the closeness of data points in high-dimensional space can be made visible. We applied t-SNE to the streamlines' feature vectors taken from the 4-th (pre-dense) network layer to examine if the streamline groups were clustered together. For the sets of streamlines, we used the pseudo ground truth and inconclusive streamlines with $0\% <= AR < 20\%$, or $80\% < AR <= 100\%$, respectively. These inconclusive streamlines are expected to be closer to the pseudo ground truth data.

\begin{figure}
\begin{center}
\begin{tabular}{c}
\includegraphics[width=0.35\textwidth]{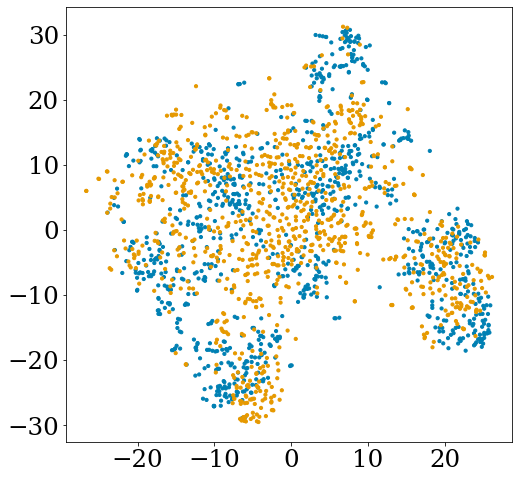} \\
a) P/N\\
\includegraphics[width=0.35\textwidth]{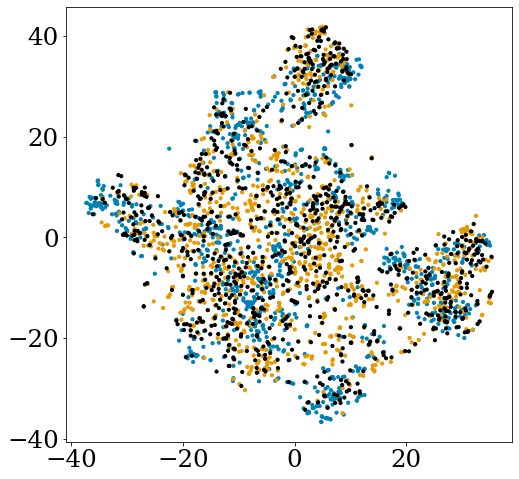} \\
  b) P/N/I($AR \leq 20\%$) \\
\includegraphics[width=0.35\textwidth]{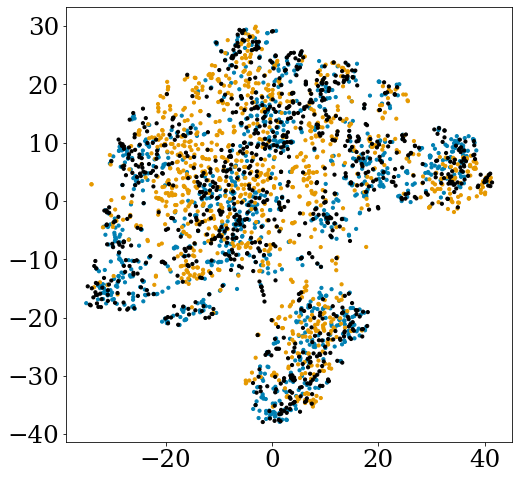} \\
  c) P/N/I($AR \geq 80\%$) 
\end{tabular}
\end{center}
\caption{t-SNE results.
Orange dots represent streamlines from the implausible pseudo ground truth, blue dots represent streamlines from the plausible pseudo ground truth, and black dots show inconclusive streamlines. The upper row does not include inconclusive samples. In the middle row, inconclusive samples with an acceptance rate AR $\leq$ 20\% in rSIFT were used. The bottom row shows the results with inconclusive streamlines with AR $\geq$ 80\%. }
\label{fig:tSNE}
\end{figure}

Plots of t-SNE were fine-tuned to a learning rate and perplexity of 100 each and 800 iterations. The comparison of t-SNE results with and without inconclusive streamlines using balanced sampling can be seen in Fig.~\ref{fig:tSNE}.

As shown, the positive and negative streamlines formed multiple small clusters which were positioned close to each other (cf. Fig. \ref{fig:tSNE}a). When additionally presented with samples from the inconclusive class, small differences were noticeable depending on if samples with low or high AR were used. Especially the latter seemed to be more close to the clusters of the plausible streamlines (cf. blue and black dots in Fig. \ref{fig:tSNE}c). 

\section{Discussion}
Tractogram filtering based on methods like SIFT offers the opportunity to improve the match of a given tractogram with the measured diffusion MRI data. However, the ability of SIFT to assign binary labels of streamline plausibility is inherently limited. In this paper, we propose a randomized iterative adaptation, called rSIFT, as a way to address this limitation of SIFT. We analyzed rSIFT in experiments on human and phantom data and employed neural network-based streamline classifiers further to characterize the properties of the output of our method. Additionally, the trained classifiers underline the potential for tractogram filtering being performed by neural networks at a reduced computational cost. In the following subsections, we discuss different aspects of our findings in further detail.

\subsection{SIFT}
Our experiments highlight how the filtering performed by SIFT depends not only on the characteristics of a streamline but also on the size and composition of the streamline set used as the input. As shown in the presented results, the same streamline can be accepted and rejected in different SIFT runs depending on the streamlines contained in the analyzed tractogram. This ambiguity, reflected in the inconclusive class from rSIFT, is more prominent with smaller tractograms. Through the randomized and iterative approach, rSIFT is designed to exploit this dependency of the SIFT output on the composition of the input tractogram.

This dependency on the input tractogram is rooted in the global optimization approach for improving the consistency between tractogram and diffusion data. It implies that the focus of SIFT lies in fitting all streamlines simultaneously to the diffusion data instead of assessing an individual streamline alone. As shown in the experiments, filtering a tractogram containing fewer streamlines with SIFT will directly lead to an increase in accepted streamlines (2.6\% for the whole tractogram vs. 14.9\% for the smallest subset size), a decrease in rejected ones (96.6\% vs. 56.8\%, respectively), and an increase in streamlines for which SIFT gives mixed results (0.7\% vs. 28.4\%). One possible explanation is that smaller tractograms might contain fewer redundant streamlines making it more likely for streamlines to be accepted.

As a consequence, we argue that SIFT in its original form with only a single run is not perfectly suited for the assessment of individual streamline plausibility with respect to the diffusion data. In fact, for some of the HCP tractograms, the results differed even when the set of streamlines did not: In the experiment instance where the complete tractogram was repeatedly filtered (i.e. subset size 10 million), around 0.7\% of the streamlines received mixed votes even though the overall tractogram remained the same, but was arranged in shuffled order due to our random sampling of streamlines.  

A general limitation of SIFT is that the method is indifferent to \textit{why} streamlines do not match the data. As mentioned in the introduction, anatomical implausibility is just one possible reason, but another possible explanation can be that the streamline in question contributes to a white matter fiber bundle that has been reconstructed with exaggerated streamline density compared to other bundles, relative to the measured data. This is supported by our findings on the DiSCo data: As shown in Fig.
~\ref{fig:disco_red}, redundant streamlines are rejected more often when they appear with a higher frequency in the dataset. At the same time, SIFT rejects true positives more often when there are no false positives present in the tractogram. Together, this shows that SIFT rejects not only implausible, but also both plausible and redundant streamlines. However, for purposes such as finding training samples for a machine learning classifier or for the purpose of combining several filtering methodologies in an ensemble fashion, as showcased in \cite{Joergens:2022}, there is a need for methods performing the distinction between plausible and implausible streamlines more reliably than what is achievable with a single run of SIFT.

\subsection{Randomized SIFT}

The proposed method, rSIFT, is a randomized, iterative adaptation of SIFT with the aim to separate plausible (including redundant) streamlines from implausible ones. Through the repeated application of SIFT to random tractogram subsets, we defined distilled sets of streamlines that are consistently categorized as either fitting the diffusion data well or not, thus deemed plausible or implausible in the sense of SIFT. Despite a large amount of votes per streamline (35 on average), the repeated SIFT assessments were consistent for around half of the streamlines in the HCP tractograms. Based on this, these streamlines might also possess distinct characteristics that a machine learning classifier may be able to learn. Thus, they are suitable for creating a pseudo ground truth of plausible and implausible streamlines with consistent characteristics within each of the two label groups, as we did in this study. 

On the other hand, around half of the streamlines were inconclusive, with ARs between 0\% and 100\%. 
Some of these streamlines may be outliers that received mostly consistent votes. Apart from those, we suspected that the group of inconclusives contains a large number of redundant streamlines as well. By choosing random subsets of the tractogram and thus omitting potentially similar streamlines that would ``rival'' a candidate redundant streamline, our filtering procedure allows such streamlines to receive more positive votes than in the context of the full tractogram. Indeed, our experiments show that this group of inconclusive streamlines likely contains a mixture of both redundant, i.e., actually plausible, as well as implausible streamlines. Naturally, alternative choices of AR thresholds for the separation of plausible, inconclusive, and implausible streamlines are possible. Actually, our results show that streamlines in the inconclusive group with higher and lower AR (e.g., higher than 80\% and lower than 20\%) share similar characteristics with plausible and implausible streamlines, respectively. This is a promising avenue for future research to better understand the composition of the group of inconclusive streamlines.

In general, it needs to be noted that the labeling of a streamline as ``plausible'' through SIFT represents more of an absence of the label ``implausible''. One could even argue that those plausible streamlines had remained in the tractogram because there were other streamlines that demonstrated to be even \textit{less} plausible before filtering was terminated. However, we think this to be unlikely as, from our experience, SIFT filters streamlines quite strictly compared to other tractogram filtering methods, leading to the removal of a significant amount of streamlines. For example, it is interesting that SIFT only kept 53\% of the streamlines in the original DiSCo data. This value was even lower for AR=100\% after rSIFT (35.0\% and 33.2\% for the false positives and redundant experiments). This means that SIFT may, in general, be too restrictive, and a threshold on AR can be set to a lower value (e.g., 80\%) to distinguish between plausible and implausible streamlines. Of course, such a threshold involves a trade-off between rejected false positives and removed true positives.

It is important to emphasize that, as discussed in Smith \textit{et al.} \cite{Smith:2015Effects}, SIFT results tend to correlate with streamline length. Since rSIFT builds on SIFT, this characteristic is inherited and also rSIFT, and the trained classifiers show this tendency to keep shorter streamlines. 
It would be interesting to assess if this correlation is a bias of SIFT or if the longer streamlines are more likely to be implausible. 

Finally, notice that while rSIFT is based on SIFT, they have different goals and applications. While SIFT assesses the consistency of the whole tractogram with respect to the measured data, rSIFT augments this goal with the assessment of the stability of these results and thus becomes more specific for individual streamlines.

\subsection{rSIFT neural network classifier}

We employed different neural network classifiers to investigate the inconclusive streamlines in rSIFT as well as to showcase the potential of such machine learning-based approaches for the purpose of reducing the computational burden of our method.

The classifier which was trained to tell apart the plausible and implausible samples showed good accuracy. This indicates that some distinct structural information that may separate the two sets of streamlines was recognized, even in data samples unseen during training.
Since rSIFT is time-consuming, it is relevant to assess whether the obtained classification scores could be used as an alternative to the SIFT-based method. The distributions of classification scores show that it is unlikely for a plausible streamline to receive a classifier score that is lower than the average (0.28) or median (0.21) scores for implausible streamlines. Considering the difference in the size of the groups of plausible and implausible streamlines, we would argue that streamlines with scores lower than approximately 0.2 could be safely removed. Such a strategy could significantly reduce the number of false-positive streamlines in a tractogram and would require little computational effort once the classification neural network has been trained.

%
Since the performance of the binary classifiers distinguishing plausible from implausible streamlines carried over to unseen data (of plausible/implausible samples), it may also be suited to identify such information in the streamlines assigned to the inconclusive group. However, employing a similar network architecture, the three-class network was unable to recognize the inconclusive samples (cf. TPR and TNR in Fig.~\ref{fig:classifiercat}). The same tendency was found for the binary approaches to distinguish inconclusive from plausible or implausible streamlines, respectively (in both cases, accuracy, sensitivity, and specificity decreased significantly compared to the N vs. P case, as seen in Fig.~\ref{fig:classifier}). This suggests that there is no inherent structural difference between inconclusive and likely plausible or implausible samples obtained through rSIFT. The findings of the t-SNE experiment further show that inconclusive streamlines are clustered together with the likely plausible and implausible samples. Additionally, their acceptance rate seems to be related to the observation of being closer to one group or the other. Taking this into account, as well as the performance of the classifiers, it can be inferred that the inconclusive samples are a mixture of plausible (containing redundant) and implausible streamlines. 

Notice that the binary classifier (plausible vs. implausible) yielded good results in both cross-validation as well as classification of unseen data despite the fact that its only input is the streamline coordinates. In contrast, SIFT considers the FOD data. This observation is consistent with the results in \cite{Joergens:2022} that showed that the coordinates of the streamlines are the most important features for the prediction of different tractogram filtering methods, followed by diffusion data.

Further, the presented classification results were achieved with a relatively simple neural network architecture. In preliminary experiments, we explored the architecture of the classifier (e.g., adding more layers or batch normalization) without a large improvement in performance. However, ways of building more sophisticated classifiers achieving even better accuracy are a direction of research that may be explored more in the future. For datasets that suffer from severe data imbalance (such as our training data), specific care may be put on recognizing the false positive samples in order to help decrease their number in comparison to the true positives.

\subsection{Future work}

There are many avenues for extending the current study. While we focused our experiments on SIFT, the same methodology can be applied to similar tractogram filtering methods such as LiFE \cite{LiFe:2014}, COMMIT \cite{COMMIT:2014}, SIFT2 \cite{Smith:2015a}, or COMMIT2 \cite{Schiavi:2020,Ocampo:2021,Sairanen:2022}. It may be particularly interesting to investigate if the presented findings would generalize to these methods as well. Also, further performance evaluation with other tractography methods and images acquired with clinical settings is relevant.

Regarding the classifier, one possibility is to add different input features as done, e.g., in \cite{TRAFIC:2018,Joergens:2022}, which described the streamline's structure in a more sophisticated manner or add diffusion data \cite{Joergens:2019,Joergens:2022}.

As discussed previously, rSIFT can be used to assess streamline plausibility or to use it for training machine learning-based classifiers, as we did in this paper. An additional interesting application would be to combine filtering and tractography in order to improve the quality of the tractogram from its generation.

\section{Conclusion}

In this paper, we proposed rSIFT, a randomized and iterative adaptation of SIFT that allows assessing the plausibility of individual streamlines with improved specificity. rSIFT was used to generate pseudo ground truths for the training of machine learning-based classifiers. These classifiers were used to study the characteristics of different types of streamlines (plausible, implausible, and inconclusive) and to speed up the computations. We show how to use AR from rSIFT or the classification scores for distinguishing plausible and implausible streamlines. Streamlines with inconclusive results from rSIFT are likely to be a mixture of redundant and implausible streamlines.

\section*{Acknowledgments}
Data for this study were provided in part by the Human Connectome Project, WU-Minn Consortium (Principal Investigators: David Van Essen and Kamil Ugurbil; 1U54MH091657) funded by the 16 NIH Institutes and Centers that support the NIH Blueprint for Neuroscience Research; and by the McDonnell Center for Systems Neuroscience at Washington University.

We thank Jakob Wasserthal for computing and providing the tractograms we used in this paper.

Antonia Hain was affiliated with Saarland University at the time the presented research was conducted and is now with the Department Quality Assurance, Characterization and Simulation at Fraunhofer Institute for Solar Energy Systems ISE, Heidenhofstr. 2, 79110 Freiburg, Baden-W\"urttemberg, Germany.

\section*{Funding sources}
This work was partially supported by Digital Futures, project dBrain. The funding sources had no involvement in the research and preparation of this article.
\section*{Author contributions}
\textbf{AH}: conceptualization; formal analysis; investigation; methodology; software; validation; visualization; writing - original draft; writing - review \& editing. 
\textbf{DJ}: conceptualization; formal analysis; investigation; methodology; software; validation; visualization; writing - original draft; writing - review \& editing.
\textbf{RM}: conceptualization; funding acquisition; methodology; visualization; resources; project administration; supervision; writing - review \& editing. 
\section*{Conflicts of interest/Competing interests}
The authors declare that they have no conflict of interest.
\section*{Data and code availability statement}
We used data from the Human Connectome Project and the DiSCo challenge. The software code used in the paper and pretrained weights of the classifier are distributed freely via the Github repository \\ https://github.com/djoerch/randomised\_filtering.

\bibliographystyle{elsarticle-num} 
\bibliography{cas-refs}

\end{document}

%% file: submission_random_res.tex

\begin{figure}
    \centering
    \includegraphics[width=\textwidth,clip,trim=0cm 1.2cm 0cm 0cm]{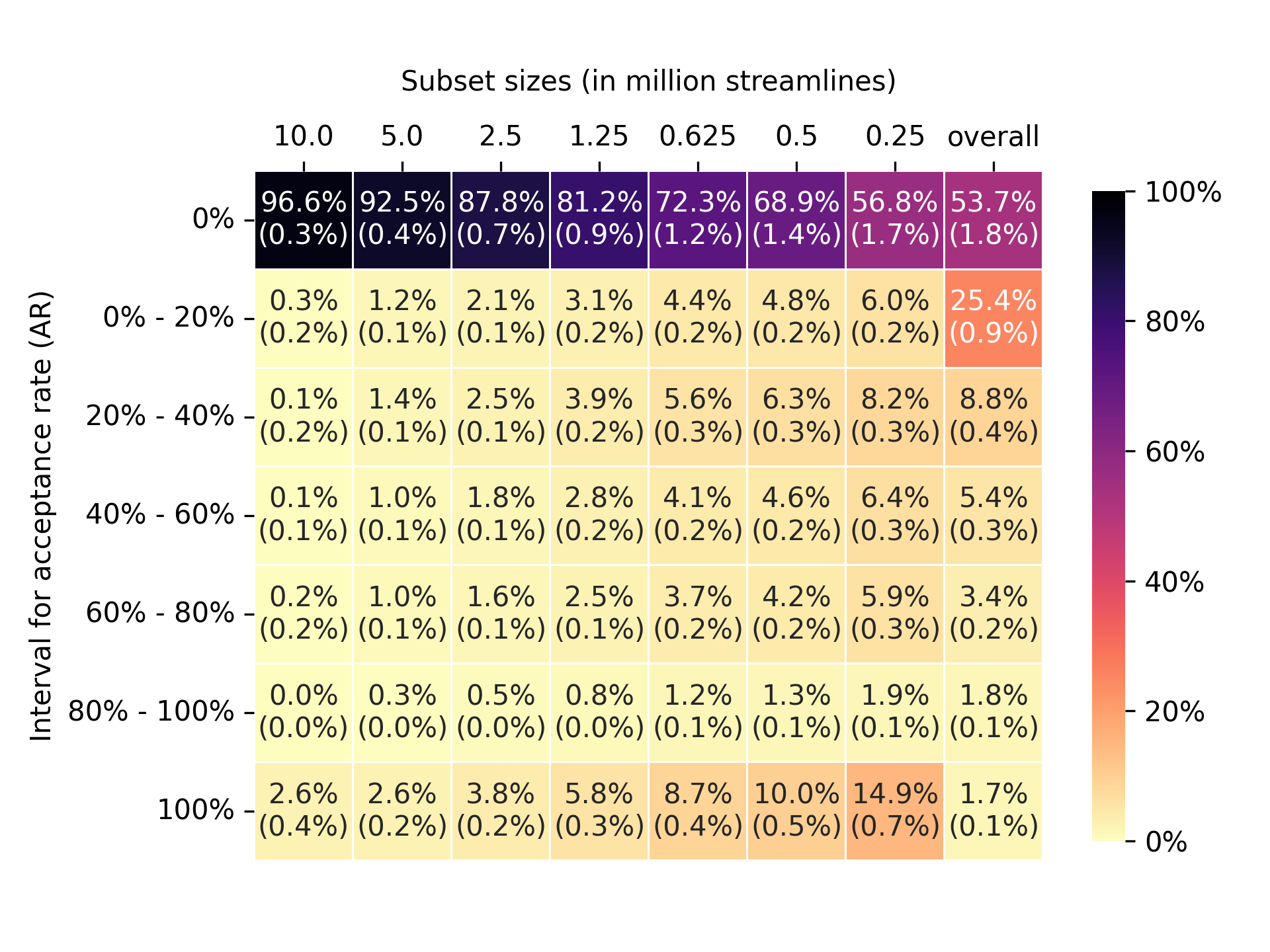}
    \caption{Distribution of streamlines per acceptance rates $AR_n$ and different subset sizes $n$. Each cell shows the mean ratio of streamlines across HCP subjects, with standard deviations in parentheses. In the last column, percentages are computed over \textit{all} experiment instances and subset sizes, thus denoting $AR$ instead of $AR_n$. Note that the AR intervals do not include the particular lower bound and include the upper bound, except for 80--100\%, where 100\% is also excluded.}
    \label{fig:table_1}
\end{figure}

\begin{figure}
    \centering
    \includegraphics[width=\textwidth,clip,trim=0cm 1.2cm 0cm 0cm]{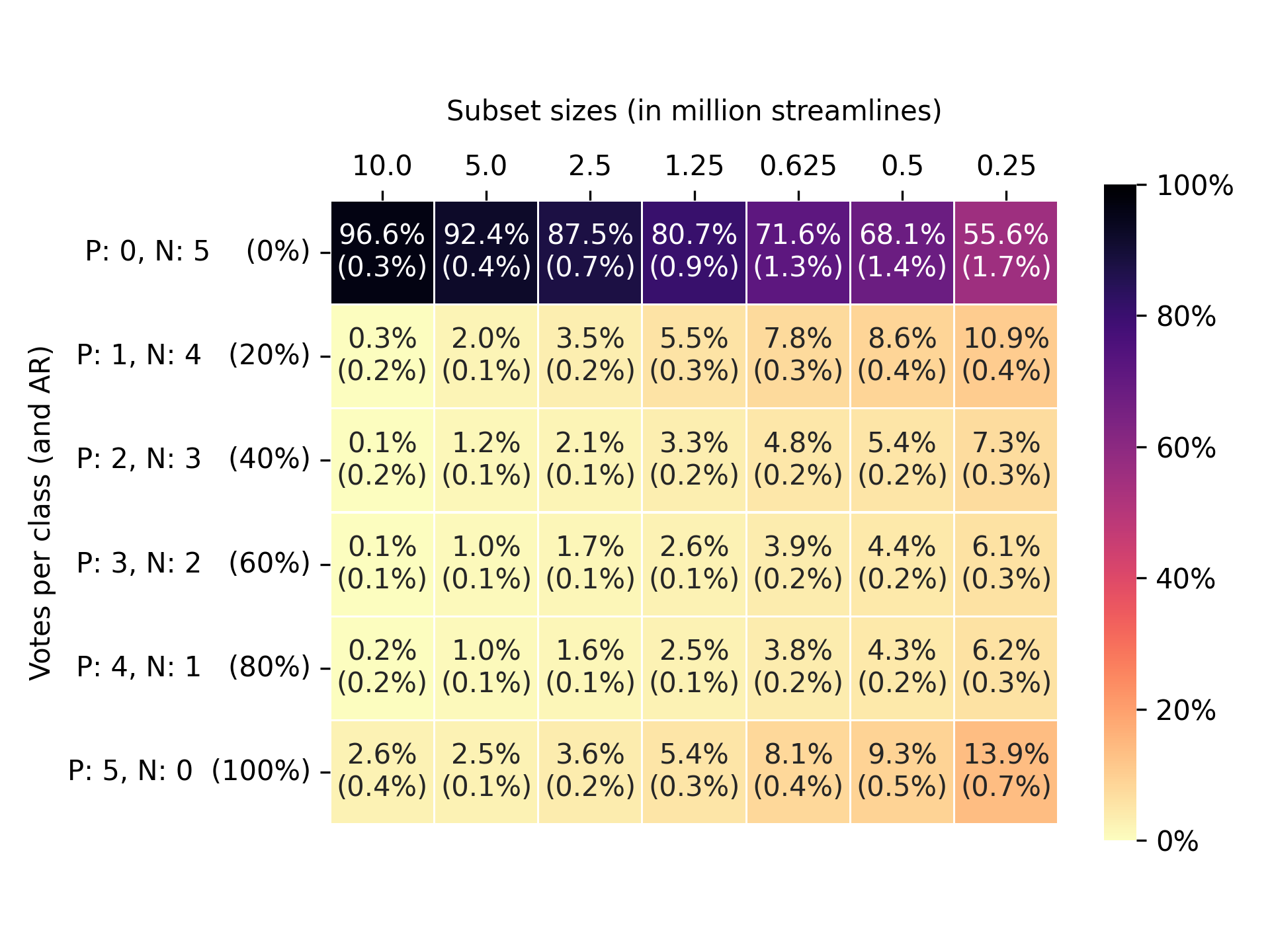}
    \caption{Distribution of streamlines per vote combination and subset size. Only combinations of exactly 5 votes are considered. ``P: 0, N: 5'' refers to ``0 positive, 5 negative votes'' with the other rows labeled analogously. The cell values show the mean ratio of streamlines across HCP subjects, with standard deviations in parentheses.}
    \label{fig:table_votesfive}
\end{figure}

\begin{figure}
\begin{tikzpicture}
		\begin{axis}[
    title={rSIFT streamline acceptance rates ($AR_n$) by subset size $n$},
    xlabel={Subset size ($\times 10^6$) },
    ylabel={Percentage of streamlines},
    xmin=0, xmax=10,
    ymin=0, ymax=100,
    xtick={0,1,2,3,4,5,6,7,8,9,10},
    ytick={0,20,40,60,80,100},
    legend style={at={(0.95,0.5)},anchor=east, 
    cells={line width=3pt}},
    legend cell align={left},
    ymajorgrids=true,
    width=\textwidth,
    height=8cm
]

\definecolor{color1}{HTML}{999999}

\addplot[color= color1,]     coordinates {(10,96.64166666666671)(5,92.47999999999999)(2.5,87.74833333333329)(1.25,81.1633333333333)(0.625,72.2883333333333)(0.5,68.895)(0.25,56.755)};
\legend{$AR_n $= 0\%}     

\addplot[forget plot,color= color1,opacity=0.2,     name path=upper1] 
coordinates {(10,96.98057176646113)(5,92.91400460827045)(2.5,88.40846715814965)(1.25,82.09247630762017)(0.625,73.21747630762017)(0.5,70.24822947056292)(0.25,58.436424990893144)};

 \addplot[forget plot,     color=color1,     opacity=0.2,     name path=lower1     ]   coordinates {(10,96.30276156687229)(5,92.04599539172953)(2.5,87.08819950851694)(1.25,80.23419035904642)(0.625,71.35919035904642)(0.5,67.54177052943707)(0.25,55.07357500910686)};

 \addplot[forget plot,color= color1,fill opacity=0.2] fill between[of=lower1 and upper1];

 \addplot[forget plot, dashed, color=color1]    coordinates {(0,53.67)(10,53.67)};
 \addplot[forget plot, dashed, color=color1, opacity=0.2, name path=lower_dashed1]    coordinates {(0,51.87)(10,51.87)};
 \addplot[forget plot, dashed, color=color1, opacity=0.2, name path=upper_dashed1]    coordinates {(0,55.47)(10,55.47)};
 \addplot[forget plot,color= color1,fill opacity=0.2] fill between[of=lower_dashed1 and upper_dashed1];

\node[color1] at (6,55) {53.67\%};

\definecolor{color2}{HTML}{E69F00}

\addplot[color= color2,]     coordinates {(10,3.36166666666667)(5,7.5249999999999995)(2.5,12.25)(1.25,18.834999999999997)(0.625,27.71)(0.5,31.1033333333333)(0.25,43.2466666666667)};     \addlegendentry{$AR_n >$ 0\%}     

     \addplot[ forget plot,    color= color2,     opacity=0.2,     name path=upper2     ] 
coordinates {(10,3.70491099169866)(5,7.956636421076812)(2.5,12.906871372492361)(1.25,19.76427391010401)(0.625,28.639273910104013)(0.5,32.45706398108664)(0.25,44.92574577935516)};

 \addplot[forget plot,     color=color2,     opacity=0.2,     name path=lower2     ]    coordinates {(10,3.01842234163468)(5,7.093363578923187)(2.5,11.593128627507639)(1.25,17.905726089895985)(0.625,26.78072608989599)(0.5,29.74960268557996)(0.25,41.567587553978235)};

 \addplot[forget plot,color= color2,fill opacity=0.2] fill between[of=lower2 and upper2];

\definecolor{color3}{HTML}{CC97A7}

\addplot[color= color3,]     coordinates {(10,3.11)(5,6.3316666666666706)(2.5,10.205)(1.25,15.715000000000002)(0.625,23.3383333333333)(0.5,26.321666666666697)(0.25,37.285000000000004)};
  \addlegendentry{$AR_n >$ 20\%}     

     \addplot[forget plot,     color= color3,     opacity=0.2,     name path=upper3     ] coordinates {(10,3.36822470834527)(5,6.693463110356582)(2.5,10.75475449065924)(1.25,16.49954445380743)(0.625,24.12287778714073)(0.5,27.497122606575406)(0.25,38.804022712140934)};

 \addplot[forget plot,     color=color3,     opacity=0.2,     name path=lower3     ]    coordinates {(10,2.8517752916547296)(5,5.969870222976759)(2.5,9.65524550934076)(1.25,14.93045554619257)(0.625,22.553788879525868)(0.5,25.14621072675799)(0.25,35.76597728785907)};

 \addplot[forget plot,color= color3,fill opacity=0.2] fill between[of=lower3 and upper3];

\definecolor{color4}{HTML}{009E73}

\addplot[color= color4,]     coordinates {(10,2.98833333333333)(5,4.90833333333333)(2.5,7.688333333333331)(1.25,11.8033333333333)(0.625,17.695)(0.5,20.064999999999998)(0.25,29.068333333333303)};   \addlegendentry{$AR_n >$ 40\%}     

     \addplot[forget plot,     color= color4,     opacity=0.2,     name path=upper4     ] 
coordinates {(10,3.168934182353113)(5,5.191205507056138)(2.5,8.102599753245284)(1.25,12.40182031453395)(0.625,18.29348698120065)(0.5,20.998354166434154)(0.25,30.321671542543672)};

 \addplot[forget plot,     color=color4,     opacity=0.2,     name path=lower4     ]    coordinates {(10,2.807732484313547)(5,4.625461159610522)(2.5,7.274066913421378)(1.25,11.204846352132652)(0.625,17.09651301879935)(0.5,19.13164583356584)(0.25,27.814995124122934)};

 \addplot[forget plot,color= color4, fill opacity=0.2] fill between[of=lower4 and upper4];

\definecolor{color5}{HTML}{F0E442}

\addplot[color= color5,]     coordinates {(10,2.86833333333333)(5,3.895)(2.5,5.92833333333333)(1.25,9.02833333333333)(0.625,13.5833333333333)(0.5,15.4483333333333)(0.25,22.7216666666667)};     \addlegendentry{$AR_n >$ 60\%}     

     \addplot[forget plot,     color= color5,     opacity=0.2,     name path=upper5     ] coordinates {(10,3.07006748182254)(5,4.118673869730016)(2.5,6.248421862753968)(1.25,9.48756820425089)(0.625,14.042568204250859)(0.5,16.183980440760948)(0.25,23.74429889120623)};

 \addplot[forget plot,     color=color5,     opacity=0.2,     name path=lower5     ]    coordinates {(10,2.66659918484412)(5,3.671326130269984)(2.5,5.608244803912692)(1.25,8.569098462415772)(0.625,13.124098462415741)(0.5,14.712686225905653)(0.25,21.699034442127168)};

 \addplot[forget plot,color= color5,fill opacity=0.2] fill between[of=lower5 and upper5];

\definecolor{color6}{HTML}{0072B2}

\addplot[color= color6,]     coordinates {(10,2.62666666666667)(5,2.92333333333333)(2.5,4.32166666666667)(1.25,6.5416666666666705)(0.625,9.87833333333333)(0.5,11.265)(0.25,16.785)};   \addlegendentry{$AR_n >$ 80\%}     

     \addplot[forget plot,     color= color6,     opacity=0.2,     name path=upper6     ] coordinates {(10,3.007752833650814)(5,3.092409253164235)(2.5,4.552354045312894)(1.25,6.876713927014619)(0.625,10.21338059368128)(0.5,11.810847964180503)(0.25,17.567681288903728)};

 \addplot[forget plot,     color=color6,     opacity=0.2,     name path=lower6     ]    coordinates {(10,2.245580499682526)(5,2.7542574135024247)(2.5,4.090979288020446)(1.25,6.206619406318722)(0.625,9.543286072985381)(0.5,10.719152035819498)(0.25,16.002318711096272)};

 \addplot[forget plot,color= color6,fill opacity=0.2] fill between[of=lower6 and upper6];

 \definecolor{color7}{HTML}{D55E00}
 
 \addplot[color= color7,]     coordinates {(10,2.62666666666667)(5,2.61166666666667)(2.5,3.8266666666666698)(1.25,5.77333333333333)(0.625,8.71666666666667)(0.5,9.945)(0.25,14.863333333333301)};     \addlegendentry{$AR_n =$ 100\%}     

     \addplot[forget plot,     color= color7,     opacity=0.2,     name path=upper7     ] coordinates {(10,3.007752833650814)(5,2.761788839135185)(2.5,4.031768267505539)(1.25,6.068674943450609)(0.625,9.012008276783948)(0.5,10.432103685060994)(0.25,15.564380169619342)};

 \addplot[forget plot,     color=color7,     opacity=0.2,     name path=lower7     ]    coordinates {(10,2.245580499682526)(5,2.461544494198155)(2.5,3.621565065827801)(1.25,5.477991723216051)(0.625,8.421325056549392)(0.5,9.457896314939006)(0.25,14.16228649704726)};

 \addplot[forget plot,color= color7,fill opacity=0.2] fill between[of=lower7 and upper7];

  \addplot[forget plot, dashed, color=color7]    coordinates {(0,1.65)(10,1.65)};
 \addplot[forget plot, dashed, color=color7, opacity=0.2, name path=lower_dashed7]    coordinates {(0,1.53)(10,1.53)};
 \addplot[forget plot, dashed, color=color7, opacity=0.2, name path=upper_dashed7]    coordinates {(0,1.77)(10,1.77)};
 \addplot[forget plot,color= color7,fill opacity=0.2] fill between[of=lower_dashed7 and upper_dashed7];
 
 \node[color7] at (5,2.8) {1.65\%};
 
\end{axis}
\end{tikzpicture}

\caption{Evolution of acceptance rates ($AR_n$) from rSIFT with the subset size $n$. Each curve shows the percentage of streamlines with the respective AR, averaged over all six HCP subjects. Standard deviations are shown in transparent (zoom in for details). Dashed lines show the respective ARs considering all subset sizes at once for AR=0\% (in grey) and AR=100\% (in orange). As shown, streamlines tend to get higher AR in smaller subset sizes.}
\label{fig:votes graph}
\end{figure}

%% file: submission.bbl
\begin{thebibliography}{10}
\expandafter\ifx\csname url\endcsname\relax
  \def\url#1{\texttt{#1}}\fi
\expandafter\ifx\csname urlprefix\endcsname\relax\def\urlprefix{URL }\fi
\expandafter\ifx\csname href\endcsname\relax
  \def\href#1#2{#2} \def\path#1{#1}\fi

\bibitem{Ugurlu:2019}
D.~Ugurlu, Z.~Firat, U.~Ture, G.~Unal, Supervised classification of white
  matter fibers based on neighborhood fiber orientation distributions using an
  ensemble of neural networks, in: Proc. MICCAI 2019: Computational Diffusion
  MRI, 2019, pp. 143--154.
\newblock \href {https://doi.org/10.1007/978-3-030-05831-9\_12}
  {\path{doi:10.1007/978-3-030-05831-9\_12}}.

\bibitem{Jeurissen:2017}
B.~Jeurissen, M.~Descoteaux, S.~Mori, A.~Leemans, Diffusion {MRI} fiber
  tractography of the brain, NMR in Biomedicine 32 (2017).
\newblock \href {https://doi.org/10.1002/nbm.3785}
  {\path{doi:10.1002/nbm.3785}}.

\bibitem{Frigo:2020}
M.~Frigo, S.~Deslauriers-Gauthier, D.~Parker, A.~A. Ould~Ismail, J.~J. Kim,
  R.~Verma, R.~Deriche, Diffusion {MRI} tractography filtering techniques
  change the topology of structural connectomes., Journal of Neural Engineering
  17~(6) (2020).
\newblock \href {https://doi.org/10.1088/1741-2552/abc29b}
  {\path{doi:10.1088/1741-2552/abc29b}}.

\bibitem{Panesar2019}
S.~S. Panesar, K.~Abhinav, F.~C. Yeh, T.~Jacquesson, M.~Collins,
  J.~Fernandez-Miranda, {Tractography for Surgical Neuro-Oncology Planning:
  Towards a Gold Standard}, Neurotherapeutics 16~(1) (2019) 36--51.
\newblock \href {https://doi.org/10.1007/S13311-018-00697-X}
  {\path{doi:10.1007/S13311-018-00697-X}}.

\bibitem{Yamada:2009}
K.~Yamada, K.~Sakai, K.~Akazawa, S.~Yuen, T.~Nishimura, {MR} tractography: A
  review of its clinical applications, Magnetic resonance in medical sciences 8
  (2009) 165--74.
\newblock \href {https://doi.org/10.2463/mrms.8.165}
  {\path{doi:10.2463/mrms.8.165}}.

\bibitem{Assaf:2017}
Y.~Assaf, H.~Johansen-Berg, M.~Thiebaut~de Schotten, The role of diffusion
  {MRI} in neuroscience, NMR in Biomedicine 32~(4) (2019) e3762.
\newblock \href {https://doi.org/10.1002/nbm.3762}
  {\path{doi:10.1002/nbm.3762}}.

\bibitem{Rheault:2020}
F.~Rheault, P.~Poulin, A.~Caron, E.~St-Onge, M.~Descoteaux, Common
  misconceptions, hidden biases and modern challenges of {dMRI} tractography,
  Journal of Neural Engineering 17 (2020).
\newblock \href {https://doi.org/10.1088/1741-2552/ab6aad}
  {\path{doi:10.1088/1741-2552/ab6aad}}.

\bibitem{Schilling:2020}
K.~G. Schilling, L.~Petit, F.~Rheault, S.~Remedios, C.~Pierpaoli, A.~Anderson,
  B.~Landman, M.~Descoteaux, Brain connections derived from diffusion {MRI}
  tractography can be highly anatomically accurate—if we know where white
  matter pathways start, where they end, and where they do not go, Brain
  Structure and Function 225 (2020) 2387–--2402.
\newblock \href {https://doi.org/10.1007/s00429-020-02129-z}
  {\path{doi:10.1007/s00429-020-02129-z}}.

\bibitem{Schilling:2021}
K.~G. Schilling, C.~M. Tax, F.~Rheault, C.~Hansen, Q.~Yang, F.-C. Yeh, L.~Cai,
  A.~W. Anderson, B.~A. Landman, Fiber tractography bundle segmentation depends
  on scanner effects, vendor effects, acquisition resolution, diffusion
  sampling scheme, diffusion sensitization, and bundle segmentation workflow,
  NeuroImage 242 (2021) 118451.
\newblock \href {https://doi.org/10.1016/j.neuroimage.2021.118451}
  {\path{doi:10.1016/j.neuroimage.2021.118451}}.

\bibitem{Maier-Hein:2016}
K.~H. Maier-Hein, P.~F. Neher, J.-C. Houde, M.-A. Côté, E.~Garyfallidis,
  J.~Zhong, M.~Chamberland, F.-C. Yeh, Y.-C. Lin, Q.~Ji, W.~E. Reddick, J.~O.
  Glass, D.~Q. Chen, Y.~Feng, C.~Gao, Y.~Wu, J.~Ma, R.~He, Q.~Li, C.-F. Westin,
  S.~Deslauriers-Gauthier, J.~O.~O. González, M.~Paquette, S.~St-Jean,
  G.~Girard, F.~Rheault, J.~Sidhu, C.~M.~W. Tax, F.~Guo, H.~Y. Mesri,
  S.~Dávid, M.~Froeling, A.~M. Heemskerk, A.~Leemans, A.~Boré, B.~Pinsard,
  C.~Bedetti, M.~Desrosiers, S.~Brambati, J.~Doyon, A.~Sarica, R.~Vasta,
  A.~Cerasa, A.~Quattrone, J.~Yeatman, A.~R. Khan, W.~Hodges, S.~Alexander,
  D.~Romascano, M.~Barakovic, A.~Auría, O.~Esteban, A.~Lemkaddem, J.-P.
  Thiran, H.~E. Cetingul, B.~L. Odry, B.~Mailhe, M.~S. Nadar, F.~Pizzagalli,
  G.~Prasad, J.~E. Villalon-Reina, J.~Galvis, P.~M. Thompson,
  F.~De~Santiago~Requejo, P.~L. Laguna, L.~M. Lacerda, R.~Barrett,
  F.~Dell’Acqua, M.~Catani, L.~Petit, E.~Caruyer, A.~Daducci, T.~B. Dyrby,
  T.~Holland-Letz, C.~C. Hilgetag, B.~Stieltjes, M.~Descoteaux, The challenge
  of mapping the human connectome based on diffusion tractography, Nature
  Communications 8~(1) (2017) 1349.
\newblock \href {https://doi.org/10.1038/s41467-017-01285-x}
  {\path{doi:10.1038/s41467-017-01285-x}}.

\bibitem{Thomas:2014}
C.~Thomas, F.~Q. Ye, M.~O. Irfanoglu, P.~Modi, K.~S. Saleem, D.~A. Leopold,
  C.~Pierpaoli, Anatomical accuracy of brain connections derived from diffusion
  {MRI} tractography is inherently limited, Proceedings of the National Academy
  of Sciences 111~(46) (2014) 16574--16579.
\newblock \href {https://doi.org/10.1073/pnas.1405672111}
  {\path{doi:10.1073/pnas.1405672111}}.

\bibitem{ExTractor:2021}
L.~Petit, F.~Rheault, M.~Descoteaux, N.~Tzourio-Mazoyer, Half of the
  streamlines built in a whole human brain tractogram is anatomically
  uninterpretable., in: Proc. OHBM, 2019, p. 1118488.
\newblock \href {https://doi.org/10.7490/f1000research.1118488.1}
  {\path{doi:10.7490/f1000research.1118488.1}}.

\bibitem{Wassermann:2016}
D.~Wassermann, N.~Makris, Y.~Rathi, M.~Shenton, R.~Kikinis, M.~Kubicki, C.-F.
  Westin, {The white matter query language: a novel approach for describing
  human white matter anatomy}, Brain Structure and Function 221~(9) (2016)
  4705--4721.
\newblock \href {https://doi.org/10.1007/s00429-015-1179-4}
  {\path{doi:10.1007/s00429-015-1179-4}}.

\bibitem{RecoBundles:2018}
E.~Garyfallidis, M.-A. Côté, F.~Rheault, J.~Sidhu, J.~Hau, L.~Petit,
  D.~Fortin, S.~Cunanne, M.~Descoteaux, Recognition of white matter bundles
  using local and global streamline-based registration and clustering,
  NeuroImage 170 (2018) 283--295.
\newblock \href {https://doi.org/10.1016/j.neuroimage.2017.07.015}
  {\path{doi:10.1016/j.neuroimage.2017.07.015}}.

\bibitem{SIFT:2012}
R.~E. Smith, J.-D. Tournier, F.~Calamante, A.~Connelly, {SIFT}:
  Spherical-deconvolution informed filtering of tractograms, NeuroImage 67
  (2013) 298--312.
\newblock \href {https://doi.org/10.1016/j.neuroimage.2012.11.049}
  {\path{doi:10.1016/j.neuroimage.2012.11.049}}.

\bibitem{Joergens:2021}
D.~J{\"o}rgens, M.~Descoteaux, R.~Moreno, Challenges for tractogram filtering,
  in: E.~{\"O}zarslan, T.~Schultz, E.~Zhang, A.~Fuster (Eds.), Anisotropy
  Across Fields and Scales, Springer International Publishing, Cham, 2021, pp.
  149--168.
\newblock \href {https://doi.org/10.1007/978-3-030-56215-1\_7}
  {\path{doi:10.1007/978-3-030-56215-1\_7}}.

\bibitem{Jbabdi:2011}
S.~Jbabdi, H.~Johansen-Berg, Tractography: Where do we go from here?, Brain
  connectivity 1 (2011) 169--83.
\newblock \href {https://doi.org/10.1089/brain.2011.0033}
  {\path{doi:10.1089/brain.2011.0033}}.

\bibitem{LiFe:2014}
F.~Pestilli, J.~Yeatman, A.~Rokem, K.~Kay, H.~Takemura, B.~Wandell, {LiFE:
  Linear Fascicle Evaluation} a new technology to study visual connectomes,
  Journal of Vision 14 (2014) 1122--1122.
\newblock \href {https://doi.org/10.1167/14.10.1122}
  {\path{doi:10.1167/14.10.1122}}.

\bibitem{COMMIT:2014}
A.~Daducci, A.~Dal~Palù, A.~Lemkaddem, J.-P. Thiran, {COMMIT}: Convex
  optimization modeling for microstructure informed tractography, IEEE
  Transactions on Medical Imaging 34~(1) (2015) 246--257.
\newblock \href {https://doi.org/10.1109/TMI.2014.2352414}
  {\path{doi:10.1109/TMI.2014.2352414}}.

\bibitem{Smith:2015a}
R.~E. Smith, J.-D. Tournier, F.~Calamante, A.~Connelly, {SIFT2: Enabling dense
  quantitative assessment of brain white matter connectivity using streamlines
  tractography}, NeuroImage 119 (2015) 338--351.
\newblock \href {https://doi.org/10.1016/j.neuroimage.2015.06.092}
  {\path{doi:10.1016/j.neuroimage.2015.06.092}}.

\bibitem{Schiavi:2020}
S.~Schiavi, M.~Ocampo-Pineda, M.~Barakovic, L.~Petit, M.~Descoteaux, J.-P.
  Thiran, A.~Daducci, A new method for accurate in vivo mapping of human brain
  connections using microstructural and anatomical information, Science
  Advances 6~(31) (2020) eaba8245.
\newblock \href {https://doi.org/10.1126/sciadv.aba8245}
  {\path{doi:10.1126/sciadv.aba8245}}.

\bibitem{Ocampo:2021}
M.~Ocampo-Pineda, S.~Schiavi, F.~Rheault, G.~Girard, L.~Petit, M.~Descoteaux,
  A.~Daducci, Hierarchical microstructure informed tractography, Brain
  Connectivity 11~(2) (2021) 75--88.
\newblock \href {https://doi.org/10.1089/brain.2020.0907}
  {\path{doi:10.1089/brain.2020.0907}}.

\bibitem{Sairanen:2022}
V.~Sairanen, M.~Ocampo-Pineda, C.~Granziera, S.~Schiavi, A.~Daducci,
  Incorporating outlier information into diffusion-weighted mri modeling for
  robust microstructural imaging and structural brain connectivity analyses,
  NeuroImage 247 (2022) 118802.
\newblock \href {https://doi.org/10.1016/j.neuroimage.2021.118802}
  {\path{doi:10.1016/j.neuroimage.2021.118802}}.

\bibitem{Tournier:2004}
J.-D. Tournier, F.~Calamante, D.~G. Gadian, A.~Connelly, Direct estimation of
  the fiber orientation density function from diffusion-weighted {MRI} data
  using spherical deconvolution, NeuroImage 23~(3) (2004) 1176--1185.
\newblock \href {https://doi.org/10.1016/j.neuroimage.2004.07.037}
  {\path{doi:10.1016/j.neuroimage.2004.07.037}}.

\bibitem{SIFT2:2015}
R.~E. Smith, J.-D. Tournier, F.~Calamante, A.~Connelly, {SIFT2}: Enabling dense
  quantitative assessment of brain white matter connectivity using streamlines
  tractography, NeuroImage 119 (2015) 338--351.
\newblock \href {https://doi.org/10.1016/j.neuroimage.2015.06.092}
  {\path{doi:10.1016/j.neuroimage.2015.06.092}}.

\bibitem{Astolfi2020}
P.~Astolfi, R.~Verhagen, L.~Petit, E.~Olivetti, J.~Masci, D.~Boscaini,
  P.~Avesani, Tractogram filtering of anatomically non-plausible fibers with
  geometric deep learning, Medical Image Computing and Computer Assisted
  Intervention – MICCAI 2020 12267 LNCS (2020) 291--301.
\newblock \href {https://doi.org/10.1007/978-3-030-59728-3\_29}
  {\path{doi:10.1007/978-3-030-59728-3\_29}}.

\bibitem{Joergens:2022}
D.~J\"orgens, P.-M. Jodoin, M.~Descoteaux, R.~Moreno, Merging multiple input
  descriptors and supervisors in a deep neural network for tractogram
  filtering, manuscript (2022).

\bibitem{Glasser:2013}
M.~F. Glasser, S.~N. Sotiropoulos, J.~A. Wilson, T.~S. Coalson, B.~Fischl,
  J.~L. Andersson, J.~Xu, S.~Jbabdi, M.~Webster, J.~R. Polimeni, D.~C. {Van
  Essen}, M.~Jenkinson, The minimal preprocessing pipelines for the {Human
  Connectome Project}, NeuroImage 80 (2013) 105--124, mapping the Connectome.
\newblock \href {https://doi.org/10.1016/j.neuroimage.2013.04.127}
  {\path{doi:10.1016/j.neuroimage.2013.04.127}}.

\bibitem{HCP:2013}
D.~C. {Van Essen}, S.~M. Smith, D.~M. Barch, T.~E. Behrens, E.~Yacoub,
  K.~Ugurbil, The {WU-Minn Human Connectome Project}: An overview, NeuroImage
  80 (2013) 62--79, mapping the Connectome.
\newblock \href {https://doi.org/10.1016/j.neuroimage.2013.05.041}
  {\path{doi:10.1016/j.neuroimage.2013.05.041}}.

\bibitem{TractSeg:2018}
J.~Wasserthal, P.~Neher, K.~H. Maier-Hein, {TractSeg} - {Fast} and accurate
  white matter tract segmentation, NeuroImage 183 (2018) 239--253.
\newblock \href {https://doi.org/10.1016/j.neuroimage.2018.07.070}
  {\path{doi:10.1016/j.neuroimage.2018.07.070}}.

\bibitem{Tournier:2010}
J.-D. Tournier, F.~Calamante, A.~Connelly, Improved probabilistic streamlines
  tractography by 2nd order integration over fibre orientation distributions,
  Proc. Intl. Soc. Mag. Reson. Med. (ISMRM) 18 (2010).

\bibitem{Smith:2012ACT}
R.~E. Smith, J.-D. Tournier, F.~Calamante, A.~Connelly,
  Anatomically-constrained tractography: Improved diffusion {MRI} streamlines
  tractography through effective use of anatomical information, NeuroImage
  62~(3) (2012) 1924--1938.
\newblock \href {https://doi.org/10.1016/j.neuroimage.2012.06.005}
  {\path{doi:10.1016/j.neuroimage.2012.06.005}}.

\bibitem{Presseau:2015}
C.~Presseau, P.-M. Jodoin, J.-C. Houde, M.~Descoteaux, A new compression format
  for fiber tracking datasets, NeuroImage 109 (2015) 73--83.
\newblock \href {https://doi.org//10.1016/j.neuroimage.2014.12.058}
  {\path{doi:/10.1016/j.neuroimage.2014.12.058}}.

\bibitem{Tournier:2019}
J.-D. Tournier, R.~Smith, D.~Raffelt, R.~Tabbara, T.~Dhollander, M.~Pietsch,
  D.~Christiaens, B.~Jeurissen, C.-H. Yeh, A.~Connelly, {MRtrix3}: A fast,
  flexible and open software framework for medical image processing and
  visualisation, NeuroImage 202 (2019) 116137.
\newblock \href {https://doi.org/10.1016/j.neuroimage.2019.116137}
  {\path{doi:10.1016/j.neuroimage.2019.116137}}.

\bibitem{Dipy:2014}
E.~Garyfallidis, M.~Brett, B.~Amirbekian, A.~Rokem, S.~Van Der~Walt,
  M.~Descoteaux, I.~Nimmo-Smith, Dipy, a library for the analysis of diffusion
  {MRI} data, Frontiers in Neuroinformatics 8 (2014) 8.
\newblock \href {https://doi.org/10.3389/fninf.2014.00008}
  {\path{doi:10.3389/fninf.2014.00008}}.

\bibitem{DiSCo:2021}
J.~Rafael-Patino, G.~Girard, R.~Truffet, M.~Pizzolato, E.~Caruyer, J.-P.
  Thiran, The diffusion-simulated connectivity (disco) dataset, Data in Brief
  38 (2021) 107429.
\newblock \href {https://doi.org//10.1016/j.dib.2021.107429}
  {\path{doi:/10.1016/j.dib.2021.107429}}.

\bibitem{Kingma:2017}
D.~P. Kingma, J.~Ba, Adam: A method for stochastic optimization (2017).
\newblock \href {http://arxiv.org/abs/1412.6980} {\path{arXiv:1412.6980}}.

\bibitem{Maaten:2008}
L.~van~der Maaten, G.~Hinton,
  \href{http://jmlr.org/papers/v9/vandermaaten08a.html}{Visualizing data using
  {t-SNE}}, Journal of Machine Learning Research 9~(86) (2008) 2579--2605.
\newline\urlprefix\url{http://jmlr.org/papers/v9/vandermaaten08a.html}

\bibitem{Smith:2015Effects}
R.~E. Smith, J.-D. Tournier, F.~Calamante, A.~Connelly, The effects of {SIFT}
  on the reproducibility and biological accuracy of the structural connectome,
  NeuroImage 104 (2015) 253--265.
\newblock \href {https://doi.org/10.1016/j.neuroimage.2014.10.004}
  {\path{doi:10.1016/j.neuroimage.2014.10.004}}.

\bibitem{TRAFIC:2018}
P.~D. Ngattai~Lam, G.~Belhomme, J.~Ferrall, B.~Patterson, M.~Styner, J.~C.
  Prieto, {TRAFIC}: Fiber tract classification using deep learning, Proceedings
  of SPIE--the International Society for Optical Engineering 10574 (2018) 37.
\newblock \href {https://doi.org/10.1117/12.2293931}
  {\path{doi:10.1117/12.2293931}}.

\bibitem{Joergens:2019}
D.~J\"orgens, P.~Poulin, R.~Moreno, P.-M. Jodoin, M.~Descoteaux, Towards a deep
  learning model for diffusion-aware tractogram filtering, in: Proc. {Int}.
  {Soc}. {Magn}. {Reson}. {Med}. {ISMRM}-{ESMRMB}, 2019, p. 3375.

\end{thebibliography}
